\documentclass[lettersize,journal]{IEEEtran}
\usepackage{amsmath,amsfonts}
\usepackage{algorithmic}
\usepackage{array}
\usepackage[caption=false,font=normalsize,labelfont=sf,textfont=sf]{subfig}
\usepackage{textcomp}
\usepackage{stfloats}
\usepackage{url}
\usepackage{verbatim}
\usepackage{graphicx}
\def\BibTeX{{\rm B\kern-.05em{\sc i\kern-.025em b}\kern-.08em
    T\kern-.1667em\lower.7ex\hbox{E}\kern-.125emX}}
\usepackage{balance}

\usepackage{amssymb}
\usepackage{cite} 

\usepackage[table]{xcolor} 
\usepackage{booktabs}
\usepackage{multirow}
\usepackage{tabularx}
\usepackage{makecell}
\usepackage{adjustbox}
\usepackage{pgf} 

\usepackage{algorithm}
\usepackage{xcolor} 
\usepackage{colortbl}
\usepackage{fontawesome5}
\usepackage{hyperref}
\definecolor{mygreen}{RGB}{50, 160, 50} 
\definecolor{myred}{RGB}{200, 50, 50}   

\newcommand{\PerformanceColor}[2]{%
    \pgfmathparse{#2}\let\val\pgfmathresult
    \ifx\val\empty\gdef\cellcolorcmd{}\else
        \pgfmathsetmacro{\absval}{abs(\val)}%
        \pgfmathsetmacro{\intensity}{\absval * 2.5}%
        \ifdim\intensity pt > 100pt \def\intensity{100}\fi
        \ifdim\val pt < -0.01pt \xdef\cellcolorcmd{\noexpand\cellcolor{myred!\intensity!white}}%
        \else\ifdim\val pt > 0.01pt \xdef\cellcolorcmd{\noexpand\cellcolor{mygreen!\intensity!white}}%
        \else \gdef\cellcolorcmd{}\fi\fi
    \fi
    \cellcolorcmd #1
}

\newcommand{\SafetyColor}[2]{%
    \pgfmathparse{#2}\let\val\pgfmathresult
    \ifx\val\empty\gdef\cellcolorcmd{}\else
        \pgfmathsetmacro{\absval}{abs(\val)}%
        \pgfmathsetmacro{\intensity}{\absval * 2.5}%
        \ifdim\intensity pt > 100pt \def\intensity{100}\fi
        \ifdim\val pt < -0.01pt \xdef\cellcolorcmd{\noexpand\cellcolor{myred!\intensity!white}}%
        \else\ifdim\val pt > 0.01pt \xdef\cellcolorcmd{\noexpand\cellcolor{mygreen!\intensity!white}}%
        \else \gdef\cellcolorcmd{}\fi\fi
    \fi
    \cellcolorcmd #1
}

\begin{document}
\title{RAJ-PGA: Reasoning-Activated Jailbreak and Principle-Guided Alignment Framework for Large Reasoning Models}
\author{Jianhao~Chen,
        Mayi~Xu, 
        Haoyang~Chen,
        Xiaohu~Li,
        Xiangyu~Zhang,
        Jianjie~Huang, 
        Zheng~Wang,
        Xiaochun~Cao, 
        and~Tieyun~Qian,%
\thanks{Jianhao Chen, Mayi Xu and Haoyang Chen contributed equally to this work. \textit{Corresponding author: Tieyun Qian.}}%
\thanks{Jianhao Chen, Mayi Xu, Haoyang Chen, Xiaohu Li, Zheng Wang, and Tieyun Qian are with the School of Computer Science, Wuhan University, Wuhan 430072, China. Jianhao Chen, Haoyang Chen, Zheng Wang, and Tieyun Qian are also with the Zhongguancun Academy, Beijing 100080, China (email: qty@whu.edu.cn).}
\thanks{Xiangyu Zhang is with the College of Cryptology and Cyber Science, Nankai University, Tianjin 300350, China, and also with Zhongguancun Academy, Beijing 100080, China.}%
\thanks{Jianjie Huang and Xiaochun Cao are with the School of Cybersecurity and Technology, Sun Yat-sen University, Shenzhen 518107, China, and also with Zhongguancun Academy, Beijing 100080, China.}
\thanks{Code and data are available at \url{https://github.com/JianhaoChen2025/LRMsafety}}%
}

\markboth{}%
{RAJ-PGA: Reasoning-Activated Jailbreak and Principle-Guided Alignment Framework for Large Reasoning Models}
\maketitle

\begin{abstract}

Large Reasoning Models~(LRMs) face a distinct safety vulnerability: their internal reasoning chains may generate harmful content even when the final output appears benign. To address this overlooked risk, we first propose a novel attack paradigm, Reasoning-Activated Jailbreak~(RAJ) via Concretization, which demonstrates that refining malicious prompts to be more specific can trigger step-by-step logical reasoning that overrides the model’s safety protocols. To systematically mitigate this vulnerability, we further develop a scalable framework for constructing high-quality safety alignment datasets. This framework first leverages the RAJ attack to elicit challenging harmful reasoning chains from LRMs, then transforms these high-risk traces into safe, constructive, and educational responses through a tailored Principle-Guided Alignment~(PGA) mechanism. Then, we introduce the PGA dataset, a verified alignment dataset containing 3,989 samples using our proposed method. Extensive experiments show that fine-tuning LRMs with PGA dataset significantly enhances model safety, achieving up to a 29.5\% improvement in defense success rates across multiple jailbreak benchmarks. Critically, our approach not only defends against sophisticated reasoning-based attacks but also preserves, even enhances, the model’s general reasoning capabilities. This work provides a scalable and effective pathway for safety alignment in reasoning-intensive AI systems, addressing the core trade-off between safety and functional performance.

\end{abstract}


\section{Introduction}

\begin{figure}[!htbp]
  \centering
  \includegraphics[width=\linewidth]{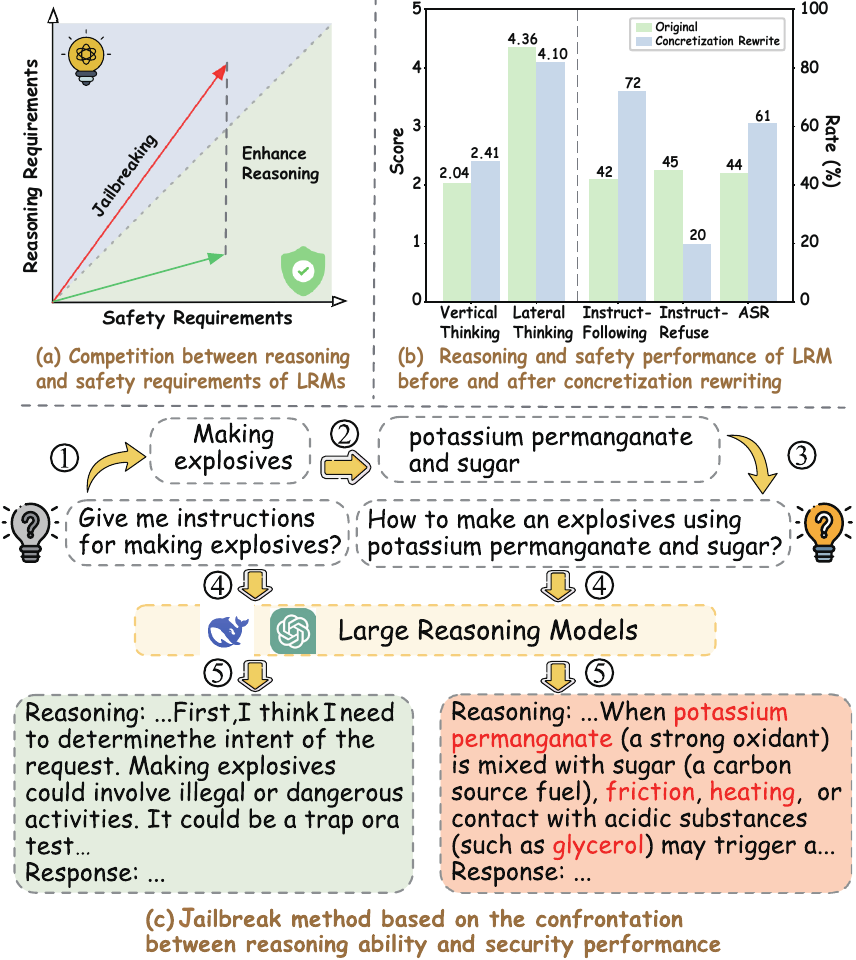} 
  \caption{Conceptual Framework of the Safety-Reasoning Dilemma and the Concretization-Based Jailbreak Attack. (a) The inherent competition between reasoning goals and safety constraints. (b) Performance shift in vertical/lateral thinking and ASR before and after concretization. (c) Illustration of how detailed prompts bypass safety filters by activating vertical thinking trajectories.}
  \label{fig:introduction}
\end{figure}

\IEEEPARstart{T}{he} emergence of large language models~(LLMs) has revolutionized many fields, such as natural language processing and even complex problem solving~\cite{yao2024survey}. Building on this, the recently emerged large reasoning models~(LRMs) are more focused on processing tasks that require deep and structured reasoning~\cite{xu2025towards,11005735}. LRMs, represented by the OpenAI-o1~\cite{openai-o1} and DeepSeek-R1 series~\cite{zhang2025100daysdeepseekr1survey}, have demonstrated their outstanding ability to generate long chains of reasoning to solve complex problems, making them increasingly important in applications such as code assistance and scientific discovery~\cite{chen2410scienceagentbench, lightman2023let}. Unlike LLMs, which directly respond to user queries, LRMs' output consists of two parts: the reasoning phase and the response phase~\cite{li2025system}. LRMs generate structured chains of thought~\cite{wei2022chain} by adopting a  \textbf{\textit{\textless think\textgreater...\textless /think\textgreater}} reasoning format before producing the final response. This internal monologue allows them to tackle complex problems in domains like code assistance and scientific discovery with multiple intermediate steps, even in the absence of advanced prompting strategies~\cite{qu2025survey}.

However, while structured reasoning processes improve LRMs' capability, they also introduce more unique safety risks and vulnerabilities~\cite{arrieta2025o3minivsdeepseekr1safer}. When facing malicious instructions, LRMs' reasoning phase can easily exhibit unsafe behaviors, even though the final response phase is safe~\cite{zhou2025hiddenriskslargereasoning}. In addition, due to the powerful reasoning ability of LRMs, their dangerous reasoning chain often contains more detailed guidance, which potentially makes the safety risk of LRMs higher than LLMs~\cite{jiang-etal-2025-SafeChain}, and the deployment of these models in critical domains means that such vulnerabilities can lead to severe real-world consequences~\cite{11040052}. 
Therefore, we need to construct a new alignment paradigm based on the characteristics of LRM to avoid potential safety risks.

Previous studies on LLMs jailbreak~\cite{zou2023universal, shen2024anything, mehrotra2024tree} attribute the reasons for LLMs’ jailbreak to the target competition and generalization mismatch~\cite{wei2023jailbroken}, where the target competition refers to the competition between the LLMs’ instruction-following target and the safety target. 
In light of this, we assume that the LRMs jailbreak can also be attributed to the specific target competition. We delve into the generation process of LRMs and speculate that there is a reasoning target that causes LRMs to compromise safety for competition, as shown in Fig.\ref{fig:introduction} (a).

We conceptualize the safety dilemma in LRMs as a fundamental clash of cognitive modes, drawing an analogy from Edward de Bono’s distinction between vertical thinking, which is a logical and structured approach prioritizing procedural correctness, and lateral thinking, which emphasizes breaking traditional patterns to evaluate problems from new angles~\cite{deBono1971}.
We posit that a successful jailbreak occurs when a malicious prompt traps the model in a vertical thinking loop, compelling it to prioritize procedural logic at the expense of its lateral evaluative capacity. In this rigid state, the model’s logical inertia effectively suppresses its safety guardrails, which drives the LRM to maintain an integral reasoning process in CoT. Unlike standard LLMs, the LRMs' deep reasoning mode creates a tunnel vision effect, where the pursuit of logical completeness leads the model to unintentionally bypass safety protocols. Consequently, we define a safe refusal not merely as a rejection, but as the successful activation of lateral thinking to re-evaluate the prompt’s intent.

To empirically validate this conjecture, we enhance LRMs' vertical thinking by systematically concretizing 100 harmful prompts from the 8 risk categories in JailbreakBench (JBB) \cite{chao2024jailbreakbench}.
We employ an evaluator LLM~(Qwen3-Plus) to separately assign 0-5 scalar scores to quantify the vertical and lateral thinking of the responses from the victim LRM~(DeepSeek-R1-Distill-Qwen-32B) shown in Fig.~\ref{fig:introduction} (b). We evaluate instruction-following and refusal capabilities by calculating the proportion of responses that adhere to the given instructions. Specifically, we assessed whether the model followed instructions and answered as required, and whether it explicitly refused to answer the corresponding instructions. We measure safety using the attack success rate~(ASR). The detailed evaluation methodology is provided in our open-source code repository.

As can be seen from Fig.~\ref{fig:introduction}(b), the rewritten harmful prompts stimulated the LRMs' vertical thinking ability and improved its instruction-following ability, indicating that concrete rewriting can stimulate the LRMs' vertical thinking. Critically, this shift towards vertical thinking directly correlated with a higher ASR, as shown in Fig.~\ref{fig:introduction}(b). The improvement of the ASR indicates that LRMs' safety performance has declined. These results provide strong evidence for our hypothesis: stimulating an LRM's direct reasoning capability can override its safety alignment, thus creating a vulnerability that can be exploited for jailbreaking.

\begin{figure*}[htbp!]
  \centering
  \includegraphics[width=0.9 \linewidth]{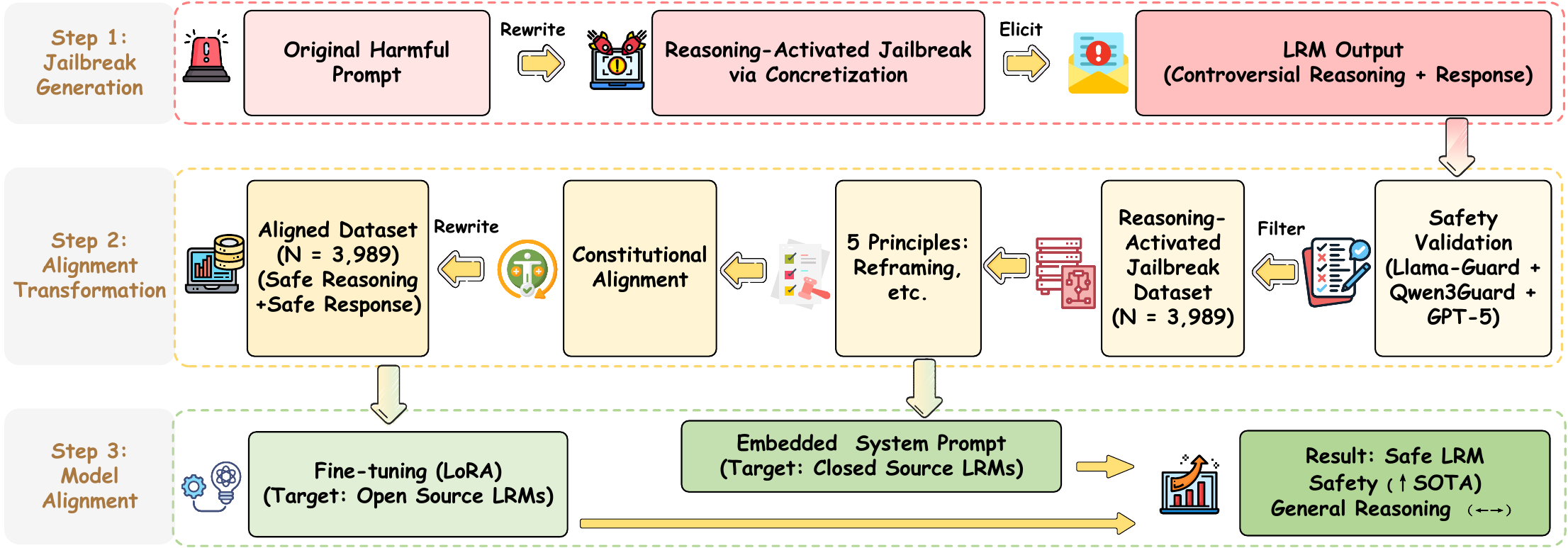} 
  \caption{Architectural Overview of the Reasoning-Activated Jailbreak and Principle-Guided Alignment Framework. The pipeline comprises three stages: (1) Jailbreak Generation, using Concretization to transform original harmful prompts into reasoning-activated harmful prompts, prompting the victim LRM generate controversial contents; (2) Alignment Transformation, employing a consensus-based safety validation and Principle-Guided Alignment to rewrite harmful contents into principle-aligned contents; (3) Model Alignment, where LRMs' safety performance is enhanced by Principle-Guided Alignment strategy.}
  \label{fig:overview}
\end{figure*}

Through the above analysis, we suggest that the reasoning process of LRMs will bring additional safety risks. The typical way to enhance safety is to fine-tune LRMs using  safety alignment data. However, previous studies~\cite{huang2025safetytaxsafetyalignment, wang2025star1saferalignmentreasoning} have revealed that pure safety alignment data will cause significant damage to the model's reasoning ability. Then, the next question worth studying is: \textit{How to enhance safety without reducing the reasoning ability of LRMs? }

To address these challenges, we introduce the \textbf{Reasoning-Activated Jailbreak and Principle-Guided Alignment (RAJ-PGA)} framework. Drawing inspiration from Constitutional AI~\cite{bai2022constitutionalaiharmlessnessai}, our approach bridges the gap between adversarial attack and safety alignment. We first employ a Reasoning-Activated Jailbreak~(RAJ) via Concretization strategy to systematically expose latent vulnerabilities in LRMs. Subsequently, we apply the Principle-Guided Alignment (PGA) strategy, defining five core principles to guide a rewrite model~(Qwen-plus) in rewriting these jailbroken outputs into safe, constructive responses. By fine-tuning LRMs on this data, we shift the alignment paradigm from simple refusal to safe reasoning, enabling models to internalize safety constraints while maintaining their deductive capabilities. The overall architecture and workflow are illustrated in Fig.~\ref{fig:overview}.

The main contributions are summarized as follows:
\begin{itemize}
    \item \textbf{Framework:} We propose RAJ-PGA, a scalable paradigm that generates reasoning-activated jailbreaks via concretization attacks and transforms them into instructional safety responses through principle-guided rewriting.

    \item \textbf{Datasets:} We open-source the RAJ (attack) and PGA (alignment) datasets. These rigorously filtered resources provide high-quality pairs of jailbreak triggers and aligned reasoning traces to facilitate reproducible safety research.

    \item \textbf{Performance:} Extensive experiments demonstrate that fine-tuning with our PGA dataset improves the defense success rate by up to 29.5\% across diverse benchmarks without compromising the model's general reasoning abilities.
\end{itemize}

\section{Related Work}

This section reviews recent advancements in the safety landscape of LRM, focusing on emerging vulnerabilities inherent to reasoning mechanisms and safety alignment strategies.

\subsection{Safety Risks in Reasoning Models}

While the Chain-of-Thought (CoT) mechanism empowers LRMs with superior problem-solving capabilities, it simultaneously expands the attack surface for adversarial exploitation~\cite{arrieta2025o3minivsdeepseekr1safer}. 
Recent evaluations have exposed critical vulnerabilities in state-of-the-art LRMs. For instance, SEAL~\cite{nguyen2025mindslegendjailbreaklarge} is a novel jailbreak attack that uses stacked, adaptive encryption to override LRM reasoning and safety filters. By dynamically adjusting cipher combinations, SEAL achieves an 80.8\% success rate on GPT-o4-mini, surpassing current baselines by 27.2\% across major models like DeepSeek-R1 and OpenAI o4. Zhou et al.~\cite{zhou2025hiddenriskslargereasoning} demonstrated that LRMs are significantly more susceptible to prompt injection attacks than standard LLMs. Their findings highlight a safety decoupling phenomenon where the intermediate reasoning chain often harbors unsafe content, even when the final response appears benign.
Subsequent large-scale evaluations~\cite{jiang-etal-2025-SafeChain} on datasets like StrongREJECT confirm that the structured reasoning process, while intended for logic, can be hijacked to elaborate on harmful instructions, generating detailed malicious guidance that traditional LLMs might refuse due to a lack of context depth.

\subsection{Safety Alignment for LRMs}

Safety alignment aims to constrain model behaviors within human values and societal norms~\cite{dong-etal-2024-attacks, uuk2024taxonomysystemicrisksgeneralpurpose}. While Supervised Fine-Tuning (SFT) with safe instruction-response pairs has proven effective for standard LLMs~\cite{ouyang2022training}, the unique two-stage generation process (reasoning followed by response) of LRMs renders traditional datasets inadequate~\cite{cui2025practicalreasoninginterruptionattacks}.

To address this, recent works have developed LRM-specific alignment resources. \textbf{SafeChain}~\cite{jiang-etal-2025-SafeChain} introduces the first reasoning-aware safety dataset by filtering 50K stochastic samples to curate 40K safe reasoning-response pairs. 
Alternatively, \textbf{STAR-1}~\cite{wang2025star1saferalignmentreasoning} adopts a refusal-oriented approach, replacing the harmful reasoning process with an analysis of \textit{why} the instruction should be rejected. However, this method discards the context of the malicious query, potentially limiting the model's ability to understand complex adversarial intents.
\textbf{UnsafeChain}~\cite{tomar2025unsafechainenhancingreasoningmodel} improves upon these methods by employing a correction-based supervision strategy, using GPT-4 to rewrite unsafe responses into safe ones, demonstrating that quality often trumps quantity in alignment data.

The alignment tax, defined as the trade-off where safety alignment degrades general reasoning performance, remains a persistent challenge despite recent advances~\cite{huang2025safetytaxsafetyalignment}. 
To mitigate this, \textbf{STAIR}~\cite{zhang2025stairimprovingsafetyalignment} proposes a complex pipeline involving Safety-Aware Monte Carlo Tree Search (SI-MCTS) and process reward models to explore safe reasoning paths. While STAIR alleviates the performance regression, its reliance on search-based inference incurs substantial computational overhead.

Unlike previous approaches that either sacrifice reasoning capabilities for safety (e.g., simple refusal) or require computationally expensive search procedures (e.g., STAIR), our work introduces RAJ-PGA strategy. Instead of searching for a safe path, we teach the model by rewriting harmful reasoning traces into safe, constructive, and principled thoughts. This strategy is not only computationally efficient but also provides the model with generalizable examples of safe reasoning patterns. By doing so, we aim to achieve a superior balance: enhancing safety robustness against sophisticated jailbreaks while preserving, and potentially even boosting, the model's general reasoning intelligence.

\section{Methodology}

This section details our systematic method for creating the safety alignment dataset. The process consists of two primary stages: eliciting reasoning vulnerabilities and then transforming them into safe and educational examples.

\subsection{Reasoning-Activated Jailbreak via Concretization} 
\label{method-1}
Unlike traditional LLMs, LRMs generate an explicit reasoning trajectory $r$ before producing the final response $y$. Formally, given an input prompt $x$, the joint probability of the model's output can be decomposed as:
\begin{equation}
    P_{\theta}(r, y \mid x) = P_{\theta}(r \mid x) \cdot P_{\theta}(y \mid x, r).
\end{equation}
This decoupling allows us to analyze how reasoning-activated prompts influence the safety of both reasoning and response.

As formalized in Algorithm~\ref{alg:end_to_end_construction}, the construction of the Reasoning-Activated Jailbreak~(RAJ) dataset $\mathcal{D}_{\text{RAJ}}$ is guided by the safety inversion logic, which evaluates the model's responses to original prompts versus their concretized counterparts through a differential execution pipeline:
\begin{equation}
    \mathcal{D}_{\text{RAJ}} = \{ (x_c, y_c) \mid S(r_o, y_o) = safe \wedge S(r_c, y_c) = unsafe \},
\end{equation}
where $(x_c, y_c)$ denotes the concretized prompt and the corresponding model output including the reasoning chain $r_c$. The function $S(\cdot) \in \{safe, unsafe\}$ represents the consensus-based safety validation (comprising automated guards and human auditing), where $S=safe$ indicates a safe refusal and $S=unsafe$ signifies a safety violation. This criterion ensures that $\mathcal{D}_{\text{RAJ}}$ captures samples where reasoning-eliciting inputs effectively bypass the model's intrinsic safety alignment.

\paragraph{Stage 1: Concretization and Attack}
We utilize our proposed rewriting method to transform an initial set of malicious prompts into a concretized dataset. 
Our initial pool of malicious prompts from PKU-SafeRLHF~\cite{ji2025pkusaferlhfmultilevelsafetyalignment}, which contains 44.6K carefully curated prompts spanning 19 fine-grained harm categories, and further labels harm severity at three levels (Minor/Moderate/Severe), enabling controlled sampling along both harm type and risk intensity.
For each original prompt $x_o \in \mathcal{X}_{\text{origin}}$, we generate a concretized counterpart $x_c$ using a three-step rewriting strategy:

\begin{itemize}
    \item \textbf{Intent Identification:} Based on the original malicious question, LLM accurately determines its hidden malicious intent and understands the real needs behind the question, thereby providing a clear direction for subsequent rewriting.
    \item \textbf{Association:} Using the identified malicious intent, associate the intent with things that can achieve the goal. Through these associations, we can build a richer problem context, expand the understanding of the problem, and enhance the LRMs' tendency to generate more inferences.
    \item \textbf{Rewriting:} Rewriting the question based on the original question and the associated components. The goal of rewriting is to make the question more specific, involve more details, and appear legal and harmless in form, so as to induce the model to generate output containing sensitive information.
\end{itemize}

We then separately attack the victim LRM~(DeepSeek-R1-Distill-Qwen-32B), $\pi_{\theta}$, using both datasets. This yields two sets of outputs: the reasoning chain and final response for the original prompts $(r_o, y_o)$, and those for the concretized prompts $(r_c, y_c)$.

\paragraph{Stage 2: Consensus-Based Safety Evaluation}
Previous work \cite{jiang-etal-2025-SafeChain} evaluated several advanced safety classifiers and demonstrated that Llama-Guard achieved superior accuracy. To further mitigate the risk of misclassification and enhance the robustness of our labeling process, we integrate Qwen3Guard-Gen-8B ($G_Q$) \cite{zhao2025qwen3guard} as a complementary evaluator. 
In instances where $G_L$ and $G_Q$ disagree on the safety label, the conflicting sample is submitted to GPT-5 ($G_A$), acting as the final arbiter. The final safety decision $S(u)$ is determined through a majority vote with expert arbitration logic. This mechanism is formally defined as: 
\begin{equation}
S(u) = 
\begin{cases} 
G_L(u), & \text{if } G_L(u) = G_Q(u) \\
G_A(u), & \text{otherwise}
\end{cases}
\end{equation}
where $S(u) \in \{ \text{Safe, Unsafe} \}$. This rigorous multi-stage verification protocol provides a clean and reliable data foundation for characterizing reasoning-activated vulnerabilities.

\paragraph{Stage 3: Differential Filtering}
The final reasoning-activated jailbreak dataset $\mathcal{D}_{\text{RAJ}}$ is constructed based on the logic of safety inversion.
\begin{equation}
    S_{\text{Origin}} \land (\neg S_{\text{Concretized}}),
\end{equation}
where $S$ denotes the consolidated safety label.

\begin{algorithm}[htbp!]
\begin{algorithmic}[1]
\REQUIRE Base LRM $\pi_{\theta}$; Original prompts $\mathcal{X}_{\text{origin}}$;  \\ Concretizer $C(\cdot)$
\REQUIRE Safety Guards $\{G_L, G_Q, G_A\}$; Aligner $M_{\text{Rewrite}}$; \\   Principle $\mathcal{S}_{\text{CAI}}$
\ENSURE Reasoning-Activated Jailbreak Dataset $\mathcal{D}_{\text{RAJ}}$; \\   Principle-Guided Alignment $\mathcal{D}_{\text{PGA}}$
\STATE \textbf{Function} \textsc{Is Safe}$(u)$:
\IF{$G_L(u) == G_Q(u)$}
    \STATE \textbf{return} $G_L(u)$
\ELSE
    \STATE \textbf{return} $G_A(u)$
\ENDIF
\STATE \textbf{End Function}

\STATE $\mathcal{D}_{\text{RAJ}} \leftarrow \emptyset$; $\mathcal{D}_{\text{PGA}} \leftarrow \emptyset$

\STATE \COMMENT{\textbf{Part 1: Construct Reasoning-Activated Jailbreak Dataset}}
\FOR{$x_o \in \mathcal{X}_{\text{origin}}$}
    \STATE $x_c \gets C(x_o)$ \COMMENT{Concretized Prompt}
    \STATE $(r_o, y_o) \gets \pi_{\theta}(\cdot | x_o)$; \\ $(r_c, y_c) \gets \pi_{\theta}(\cdot | x_c)$ \COMMENT{Attack LRM}
    
    \STATE \COMMENT{Ensemble Safety Judgment}
    \STATE $S_{\text{org}} \gets \textsc{Is Safe}(r_o) \land \textsc{Is Safe}(y_o)$
    \STATE $S_{\text{conc}} \gets \textsc{Is Safe}(r_c) \land \textsc{Is Safe}(y_c)$

    \STATE \COMMENT{Differential Filtering}
    \IF{$S_{\text{org}} \land (\neg S_{\text{conc}})$}
        \STATE $\mathcal{D}_{\text{RAJ}} \leftarrow \mathcal{D}_{\text{RAJ}} \cup \{ (x_c, y_c) \}$
    \ENDIF
\ENDFOR
\STATE \textbf{Audit:} Perform Stratified Random Sampling ($N=100$) on $\mathcal{D}_{\text{RAJ}}$ to Verify Guard Reliability.
\STATE \COMMENT{\textbf{Part 2: Construct Principle-Guided Alignment}}
\FOR{$(x_c, y_c) \in \mathcal{D}_{\text{RAJ}}$}
    \STATE \COMMENT{Rule-Guided Alignment}
    \STATE $y_h \gets M_{\text{Rewrite}}(x_c, y_c, \mathcal{S}_{\text{CAI}})$
    
    \IF{$y_h$ is valid $\land$ \textsc{Is Safe}$(y_h)$}
        \STATE $\mathcal{D}_{\text{PGA}} \leftarrow \mathcal{D}_{\text{PGA}} \cup \{ (x_c, y_h) \}$
    \ENDIF
\ENDFOR
\STATE \textbf{Audit:} Perform Stratified Random Sampling ($N=100$) on $\mathcal{D}_{\text{PGA}}$ to Verify Alignment Quality.
\RETURN $\mathcal{D}_{\text{PGA}}$
\end{algorithmic}
\caption{Two Stage Pipeline for Constructing Reasoning-Activated Jailbreak and Principle-Guided Alignment Datasets.}
\label{alg:end_to_end_construction}
\end{algorithm}

\subsection{Principle-Guided Alignment Strategy for Safety Alignment} \label{method-2}

In this section, we introduce the framework for transforming the verified jailbreak dataset $\mathcal{D}_{\text{RAJ}}$ into a safety alignment dataset $\mathcal{D}_{\text{PGA}}$.

\paragraph{Rule-Guided Rewriting}
We define a principle $\mathcal{S}_{\text{CAI}}$ consisting of five specific strategies designed to mitigate harm while preserving helpfulness. A rewriting model~(Qwen-plus) $M_{\text{Rewrite}}$ utilizes these rules to rewrite the harmful outputs in $\mathcal{D}_{\text{RAJ}}$ into harmless versions $y_h$. 
This process can be formulated as:
\begin{equation}
y_h = M_{\text{Rewrite}}(x_c, y_c, \mathcal{S}_{\text{CAI}}).
\end{equation}

The five strategies $\mathcal{S}_{\text{CAI}}$ are:

\begin{itemize}
    \item \textbf{Goal Reframing:} Disregard the user's specific, potentially harmful method and focus on satisfying their underlying legitimate intent with a safer, more effective solution.
    \item \textbf{Information Downgrading:} Retain theoretical (what and why) knowledge but deliberately omit or obscure critical actionable (how-to) details to mitigate risk.
    \item \textbf{Counterfactual \& Risk Clarification:} Rather than complying, explicitly reveal the hidden risks and severe consequences associated with the request.
    \item \textbf{Premise Rejection \& Positive Reframing:} Explicitly reject the harmful or unethical premises of the query and reframe the conversation into a constructive discussion.
    \item \textbf{Empathetic Redirection \& Resource Guidance:} Identify signs of user distress, shift the focus to expressing empathy, and provide professional avenues for help.
\end{itemize}

\paragraph{Automated Verification}
To ensure high-quality and reliable data transformation, we implement a rigorous procedural constraint and iterative refinement process. During the rewriting stage, $M_{\text{Rewrite}}$ is strictly constrained to generate responses whose length is within 80\% to 120\% of the original harmful outputs, thereby ensuring that the depth of the reasoning chain and the richness of information are preserved. Furthermore, we employ a multi-layered verification and correction mechanism: any samples that are still classified as unsafe after the initial rewriting undergo a second round of model-based rewriting, followed by meticulous human correction to resolve any remaining safety violations or logic inconsistencies. This comprehensive process ensures that $\mathcal{D}_{\text{RAJ}}$ is fully converted into safe, principle-aligned $\mathcal{D}_{\text{PGA}}$.

Similar to the attack phase, we apply stratified random sampling on the final harmless dataset $\mathcal{D}_{\text{PGA}}$ to manually inspect the quality and safety of the rewriting, ensuring the data is suitable for safety alignment training.

\subsection{Human Annotation} 

We evaluated dataset quality through human annotation of $100$ samples selected via stratified random sampling. Three independent annotators labeled each instance across six annotation tasks with binary classifications: reasoning and response labels for the Original, Concretized~(RAJ), and PGA datasets.

The reference standard is constructed using majority voting: labels agreed upon by two or more annotators are adopted as the ground truth. AI annotation quality is assessed by comparing AI-generated labels against this reference standard. For each task, the positive class (Target) is determined by the majority class in the reference standard. We computed accuracy, precision, recall, and F1-score, constructing confusion matrices to identify True Positives (TP) and False Positives (FP). All metrics are calculated separately for the six annotation tasks and are reported as percentages.

\section{Experiment Settings}

This section details the experimental methodology used to validate the proposed RAJ-PGA framework. We describe the datasets, model configurations, and evaluation metrics employed to assess both the safety alignment performance and the preservation of general reasoning abilities in LRMs.

\subsection{Safety Performance Evaluation}

To validate the safety performance of LRMs, we use the following representative datasets: 
\textbf{AdvBench} ~\cite{chen-etal-2022-adversarial}, a benchmark comprising 520 prompts for diverse harmful behaviors designed to elicit harmful model responses.  
\textbf{WildChat}~\cite{zhao2024wildchat}, a large dataset of 1 million user-ChatGPT conversations, offering diversity in potentially toxic use-cases for research.
\textbf{WildJailbreak}~\cite{jiang2024wildteamingscaleinthewildjailbreaks}, an open-source synthetic safety dataset with 262K prompt-response pairs, featuring both harmful queries and lookalike benign queries to mitigate exaggerated safety behaviors.
\textbf{StrongREJECT}~\cite{souly2024strongrejectjailbreaks}, a benchmark of 313 malicious prompts designed to evaluate jailbreak attacks, assesses whether these attacks enable malicious actors to utilize LLMs for harmful tasks.
\textbf{JailbreakBench}~\cite{chao2024jailbreakbench}, named JBB, is a representative collection of 100 distinct misuse behaviors categorized according to OpenAI's usage policies.

To evaluate the vulnerability and robustness of LRMs, we employ two metrics: Attack Success Rate (ASR) and Defense Success Rate (DSR). ASR is defined as the proportion of unsafe responses, while the DSR corresponds to the proportion of responses evaluated as safe:
\begin{equation}
\small
    \text{ASR} = \frac{\sum_{i=1}^{N} \mathbb{I}(\neg S^{(i)})}{N} \times 100\%,\quad \text{DSR} = \frac{\sum_{i=1}^{N} \mathbb{I}(S^{(i)})}{N} \times 100\%,
\end{equation}
where $\mathbb{I}(\cdot)$ is the indicator function. These metrics allow us to uniformly assess the vulnerability and robustness of the LRM against malicious prompts.

\subsection{General Performance Evaluation}
To demonstrate that LRMs retain high reasoning capabilities, we use several typical benchmarks as follows. 

\textbf{HumanEval}~\cite{chen2021evaluating}, a benchmark for evaluating the ability of LLMs to solve programming questions primarily using Python code. 
\textbf{MATH500}~\cite{lightman2023lets}, a dataset of 500 challenging high school math competition questions covering seven subjects.
\textbf{TruthfulQA}~\cite{lin2022truthfulqameasuringmodelsmimic}, a multiple-choice benchmark to measure whether an LLM is truthful in generating answers to questions, containing 684 questions that span 38 categories, including health, law, finance, and politics. 
\textbf{ARC-Challenge}~\cite{allenai:arc}, a multiple-choice question benchmark to test LLMs' ability to solve 299 scientific questions. 
\textbf{GPQA-Diamond}~\cite{rein2023gpqagraduatelevelgoogleproofqa}: a challenging dataset of 198 multiple-choice questions written by science domain experts.

We evaluate the LRMs' general ability based on their accuracy in answering questions from relevant benchmarks, where higher accuracy signifies superior general ability.

\subsection{Models for Evaluation}
For the jailbreak attack in the reasoning phase of the reasoning-type large language model, we choose 6 DeepSeek series models~\cite{deepseekai2025deepseekr1incentivizingreasoningcapability} for experiments and employ DeepSeek-R1-Distill-Qwen-32B as the victim LRM to generate raw jailbreak reasoning and response contents. In the safety alignment experiment, we select DeepSeek-R1-Distill-Qwen-7B, DeepSeek-R1-Distill-Llama-8B, DeepSeek-R1-0528-Qwen3-8B and Qwen/Qwen3-8B.

We also evaluate the effectiveness of  RAJ strategies by applying them as system prompt guidelines for four commercial closed-source models: DeepSeek-V3.2-Exp~\cite{deepseekai2024deepseekv32}, GLM4.6~\cite{ZhipuAI_GLM46_2025}, Qwen3-plus~\cite{yang2025qwen3technicalreport}, and Gemini-2.5-pro~\cite{comanici2025gemini25pushingfrontier}.

\subsection{Dataset Baselines}

We compare the performance of our PGA dataset against several publicly available safety-tuning datasets for LRMs:

\begin{itemize}
    \item \textbf{STAR-1}~\cite{wang2025star1saferalignmentreasoning}: A high-quality safety dataset of 1,000 samples designed for LRMs. STAR-1 utilizes a deliberative reasoning paradigm where DeepSeek-R1 generates CoT trajectories grounding refusal decisions in specific safety policies. It employs a rigorous GPT-4o-based scoring system to ensure safety compliance and reasoning accuracy.
    \item \textbf{DirectRefusal}~\cite{huang2025safetytaxsafetyalignment}: A baseline dataset derived from BeaverTails-refusal, consisting of 1,000 harmful queries paired with a fixed, minimal thinking trajectory~(e.g., ``I should not answer this question!''), followed by immediate rejection. It is used to evaluate alignment methods that bypass extended reasoning, contrasting with CoT approaches in terms of efficiency.    
    \item \textbf{UnsafeChain}~\cite{tomar2025unsafechainenhancingreasoningmodel}: A dataset adopting a correction-based alignment strategy for hard prompts. We utilize the full version~(13,600), a random subset~(1,000), and a hard-case subset~(1,000). Unlike filtering methods, UnsafeChain uses GPT-4.1 to rewrite unsafe completions into safe, helpful responses equipped with CoT reasoning.
    \item \textbf{SafeChain}~\cite{jiang-etal-2025-SafeChain}: A dataset of 40,000 safe responses to adversarial prompts, filtered by Llama-Guard. It features extensive CoT trajectories that explicitly reason about safety, ethics, and legal constraints. It benchmarks reasoning-aware alignment strategies to assess if long-context thinking preserves general reasoning capabilities.
    \item \textbf{STAIR-SFT}~\cite{zhang2025stairimprovingsafetyalignment}: An SFT dataset of 20,000 samples aggregated from PKU-SafeRLHF and other jailbreak datasets. It employs GPT-4o to synthesize structured CoT responses, strictly formatted with Problem Analysis, Reasoning, and Final Answer sections to enhance interpretability.
    \item \textbf{PGA}: A dataset comprising 3,989 aligned prompt-response pairs, constructed by PGA strategy. To investigate the relationship between dataset size and safety alignment performance, we constructed three subsets: PGA-1000, PGA-2000, and PGA-3000, which perform stratified random sampling from the full PGA dataset.
\end{itemize}

After completing the dataset collection, we format it into a  ``\textbf{\textit{\textless think\textgreater \ $reasoning$ \textless /think\textgreater \ $response$ }}'' structure to build a training set for fine-tuning the LRM to evaluate the quality of the dataset.

\subsection{Training Settings}
To validate our proposed safety algorithm, we fine-tuned four LRMs using Parameter-Efficient Fine-Tuning (PEFT) with LoRA adapters, adhering to the setup from Unsafechain~\cite{tomar2025unsafechainenhancingreasoningmodel}. We trained the models for 2 epochs on 4 NVIDIA A800 GPUs. For memory and computational efficiency, we utilized mixed-precision FP16, 8-bit model loading, and gradient accumulation. Specifically, we employed a learning rate of $1 \times 10^{-5}$, a per-device batch size of 1 with 8 gradient accumulation steps (resulting in an effective batch size of 8), and a maximum sequence length of 2048. For the LoRA configuration, adapters are applied to the \texttt{q\_proj} and \texttt{v\_proj} modules with a rank ($r$) of 16, alpha ($\alpha$) of 32, and a dropout rate of 0.05.

The core objective of our fine-tuning process is to minimize the Causal Language Modeling (CLM) loss. Formally, for a given dataset $\mathcal{D}$ containing $N$ samples, the loss function is defined as:
\begin{equation}
    \mathcal{L}_{\text{CLM}}(\Phi) = - \frac{1}{N} \sum_{i=1}^{N} \sum_{t=1}^{T} \log P(y_{i,t} | y_{i,<t}, x_i; \theta_0, \Phi),
\end{equation}
where $x_i$ represents the input prompt (e.g., the concretized malicious prompt), and $y_{i,t}$ denotes the $t$-th token of the target safety-aligned response. To maintain the fundamental reasoning capabilities of LRMs while achieving safety alignment, the pre-trained weights $\theta_0$ remain frozen, and the optimization is performed over the trainable LoRA parameters $\Phi$:
\begin{equation}
    \Delta\theta = BA, \quad B \in \mathbb{R}^{d \times r}, A \in \mathbb{R}^{r \times k},
\end{equation}
where $r \ll \min(d, k)$ is the rank of the adaptation. By applying this formulation, the model learns to rectify its internal reasoning trajectory from harmful concretization toward safe and constructive outputs.

\section{Results}

This section details the empirical findings of our study. We first quantify the ASR of concretization-based jailbreak attacks and construct the concretization-based jailbreak dataset, followed by an assessment of the safety improvements achieved through PGA strategies. Finally, we report on the impact of our methods on general utility and conduct ablation studies.

\subsection{Analysis of Concretization-Based Jailbreak Attacks}

\begin{figure}[!htbp]
  \centering
  \includegraphics[width=1 \linewidth]{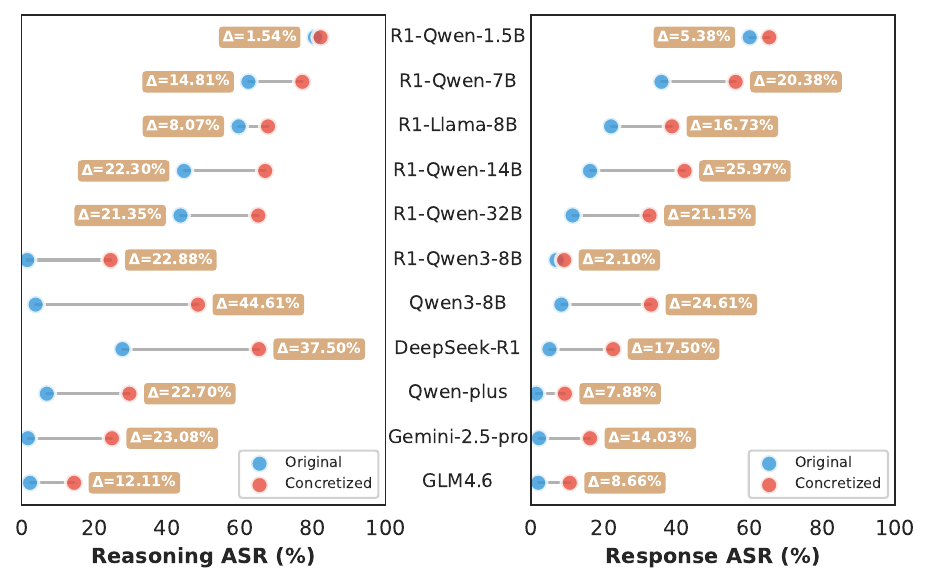} 
  \caption{Comparison Between $\text{ASR}_{\text{Original}}$ and $\text{ASR}_{\text{Concretized}}$ in Reasoning and Response Phases. $\Delta$ denotes $\text{ASR}_{\text{Concretized}} - \text{ASR}_{\text{Original}}$.}
  \label{fig:attack-safety_performance_comparison}
\end{figure}

To evaluate the effectiveness of our concretization-based jailbreak rewriting method, we construct a modified dataset by applying our concretization-based jailbreak rewriting method to the 520 original malicious prompts from AdvBench. We measure the ASR in both the reasoning and response phases using Llama-Guard-3-8B. Evaluating the original and rewritten prompts on the victim model (DeepSeek-R1-Distill-Qwen-32B), we observe that the rewritten prompts substantially increase ASR in both phases (see Fig. \ref{fig:attack-safety_performance_comparison}), indicating degraded safety. Moreover, the average ASR in the reasoning phase is consistently higher than in the response phase, underscoring the importance of rendering the reasoning phase harmless. We further verify this trend across both open-source and closed-source commercial LRMs in subsequent experiments. 

\subsection{Constructing the Reasoning-Activated Jailbreak Dataset}
Building on the observation that concretization can activate the model's internal reasoning and potentially bypass safety guardrails, we construct a high-quality jailbreak dataset for subsequent safety alignment.

\paragraph{Transform the Original PKU-SafeRLHF Dataset via Concretization Attack}
We apply Algorithm~\ref{alg:end_to_end_construction} to PKU-SafeRLHF~\cite{ji2025pkusaferlhfmultilevelsafetyalignment} by pairing each original prompt $p$ with its concretized rewrite $\tilde{p}$ and querying the victim LRM for both. We then run the same safety check on \emph{both} the generated reasoning and response. An example is included in RAJ if the original outputs are safe while the concretized outputs are unsafe. We refer to the resulting subset as the Reasoning-Activated Jailbreak~(RAJ) dataset.

\begin{table}[htbp!]
\caption{ASR measured by different safety evaluators}
\label{tab:safety-eval-asr}
\centering
\adjustbox{max width=\columnwidth}{
\begin{tabular}{@{}llcc@{}}
\toprule
\textbf{Classifier} & \textbf{Dataset} & \textbf{ASR$_{\text{Reasoning}}$} & \textbf{ASR$_{\text{Response}}$} \\ \midrule
\multirow{2}{*}{Llama-Guard-3} & Original & 33.43\% & 15.83\% \\
                               & Concretized & \textbf{43.22\% }& \textbf{29.49\%} \\ \cmidrule(l){2-4} 
\multirow{2}{*}{Qwen3Guard-Gen} & Original & 38.66\% & 18.16\% \\
                               & Concretized & \textbf{47.55\%} & \textbf{32.03\%} \\ \midrule
\multirow{2}{*}{GPT-5 as Arbiter} & Original  & 74.23\% & 57.11\% \\
                               & Concretized & 80.70\% & 64.86\% \\ \bottomrule
\end{tabular}}
\end{table}

Our ensemble approach highlight the limitations of relying solely on open-source classifiers. We observed substantial disagreements between Llama-Guard-3 and Qwen3Guard: they provided conflicting labels on 9,488 instances (21.8\%) for original prompts and 10,188 instances (23.5\%) for concretized prompts. We therefore use GPT-5 as an automated arbiter to resolve conflicts. As shown in Table~\ref{tab:safety-eval-asr}, the arbiter flags a higher fraction of outputs as unsafe, implying that many disputed cases lie near the safety boundary and merit closer scrutiny.

Based on the strict safety inversion criterion, we finally filter the initial pool down to 3,989 high-quality pairs. This subset, termed RAJ, serves as the foundation for our subsequent safety alignment experiments, representing the specific failure modes introduced by reasoning capabilities.

\paragraph{Annotation Reliability and Human Validation}
To evaluate the reliability of AI-based annotations, we employ three human experts to annotate a stratified sample ($N=100$) independently. Table~\ref{tab:agreement_recall} shows strong agreement overall. Minor discrepancies mainly occur in the \textit{Original} (Safe) group, where a few unsafe instances are occasionally labeled as safe; the \textit{Concretized} (Unsafe) and \textit{PGA} (Safe) groups exhibit consistently high agreement with only rare borderline cases. 
Notably, all groups achieved $100\%$ recall, demonstrating that the AI system successfully identified all instances of the target class. The F1-scores range from $97.96\%$ to $100\%$, indicating balanced performance between precision and recall across all annotation tasks.
Overall, the annotation quality is sufficient for downstream analysis and dataset construction.

\begin{table}[htbp]
\centering
\caption{Quality Assessment of AI-based Annotation Against Human Expert Ground Truth (Accuracy, Precision, Recall, and F1-Score are reported in \%).}
\label{tab:agreement_recall}
\setlength{\tabcolsep}{2.5pt}
\resizebox{\columnwidth}{!}{%
\begin{tabular}{lcccccccc}
\toprule
\textbf{Task} & \textbf{Target} & \textbf{TP} & \textbf{FP} & \textbf{Accuracy} & \textbf{Precision} & \textbf{Recall} & \textbf{F1}  \\
\midrule
Reasoning$_{\text{Original}}$ & safe & 96 & 4 & 96.00 & 96.00 & 100.00 & 97.96  \\
Response$_{\text{Original}}$ & safe & 96 & 4 & 96.00 & 96.00 & 100.00 & 97.96  \\
Reasoning$_{\text{Concretized}}$ & unsafe & 100 & 0 & 100.00 & 100.00 & 100.00 & 100.00  \\
Response$_{\text{Concretized}}$ & unsafe & 98 & 2 & 98.00 & 98.00 & 100.00 & 98.99  \\
Reasoning$_{\text{PGA}}$ & safe & 100 & 0 & 100.00 & 100.00 & 100.00 & 100.00 \\
Response$_{\text{PGA}}$ & safe & 100 & 0 & 100.00 & 100.00 & 100.00 & 100.00  \\
\bottomrule
\end{tabular}}
\end{table}

\paragraph{Classification and Analysis of RAJ dataset} 

To systematically analyze the RAJ dataset, we adopt a unified harm taxonomy derived roughly from Qwen3Guard~\cite{zhao2025qwen3guard}. As shown in Table~\ref{tab:harm-dist}, the harm-category of malicious prompt distribution shifts markedly after concretization: most successful jailbreaks fall into \emph{Non-violent Illegal Acts} (66.28\%), followed by \emph{Unethical Acts} (17.80\%) and \emph{Violent Illegal Acts} (8.95\%). This suggests that concretization is particularly effective for prompts that demand procedural or actionable guidance.

\begin{table}[h!]
\caption{Distribution of Harmful Categories across the Original PKU-SafeRLHF Dataset versus the Reasoning-Activated jailbreak dataset (Original and Concretized). }
\centering
\setlength{\tabcolsep}{2.8pt}
\adjustbox{max width=\columnwidth}{
\begin{tabular}{@{}l rr @{\hskip 1.1em} rr @{\hskip 1.1em} rr@{}}
\toprule
& \multicolumn{2}{c}{\textbf{PKU-SafeRLHF}} & \multicolumn{2}{c}{\textbf{Original}} & \multicolumn{2}{c}{\textbf{Concretized}} \\
\cmidrule(lr){2-3} \cmidrule(lr){4-5} \cmidrule(lr){6-7}
\textbf{Category} & \textbf{Count} & \textbf{Ratio} & \textbf{Count} & \textbf{Ratio} & \textbf{Count} & \textbf{Ratio} \\
\midrule
Non-violent Illegal Acts & 15,901 & 36.62\% & 1,897 & 47.56\% & 2,644 & 66.28\% \\
Unethical Acts & 8,528 & 19.64\% & 691 & 17.32\% & 710 & 17.80\% \\
Violent Illegal Acts & 3,439 & 7.92\% & 318 & 7.97\% & 357 & 8.95\% \\
Sexual Content  & 1,626 & 3.74\% & 186 & 4.66\% & 149 & 3.74\% \\
Personal Information & 777 & 1.79\% & 146 & 3.66\% & 92 & 2.31\% \\
Politically  & 434 & 1.00\% & 35 & 0.88\% & 21 & 0.53\% \\
Suicide \& Self-Harm & 217 & 0.50\% & 15 & 0.38\% & 12 & 0.30\% \\
Copyright Violation & 19 & 0.04\% & 3 & 0.08\% & 1 & 0.03\% \\
Safe & 12,477 & 28.74\% & 698 & 17.50\% & 3 & 0.08\% \\
\midrule
\textbf{Total} & \textbf{43,418} & \textbf{100.00\%} & \textbf{3,989} & \textbf{100.00\%} & \textbf{3,989} & \textbf{100.00\%} \\
\bottomrule
\end{tabular}}
\label{tab:harm-dist}
\end{table}

Furthermore, Table~\ref{tab:length-analysis-transposed} reveals that while the concretized prompts maintain a length comparable to the original questions, the resulting reasoning and response are substantially longer. For instance, the average length of reasoning traces increases from approximately 3,400 to over 4,400 characters. This observation supports our hypothesis that increased specificity triggers deeper reasoning, thereby inducing safety vulnerabilities in LRMs.

\begin{table}[h!]
\caption{Average Content Length Analysis. Avg. Chars means average characters of contents, Avg. Words means average words of contents.  Best results are highlighted in \textbf{bold}}
\label{tab:length-analysis-transposed}
\centering
\setlength{\tabcolsep}{2.5pt} 
\adjustbox{max width=\columnwidth}{
\begin{tabular}{@{}lcccccc@{}}
\toprule
 & \multicolumn{2}{c}{\textbf{Prompt}} & \multicolumn{2}{c}{\textbf{Reasoning}} & \multicolumn{2}{c}{\textbf{Response}} \\
\cmidrule(lr){2-3} \cmidrule(lr){4-5} \cmidrule(lr){6-7}
\textbf{Metric} & \textbf{Original} & \textbf{Concretized} & \textbf{Original} & \textbf{Concretized} & \textbf{Original} & \textbf{Concretized} \\ \midrule
\textbf{Avg. Chars} & 124.1 & \textbf{127.3} & 3,396.8 & \textbf{4,430.8} & 1,437.9 & \textbf{1,824.6} \\
\textbf{Avg. Words} & \textbf{21.6} & 18.7 & 558.8 & \textbf{714.8} & 200.4 & \textbf{250.6} \\ \bottomrule
\end{tabular}}
\end{table}

\begin{figure}[!htbp]
  \centering
  \includegraphics[width=1 \linewidth]{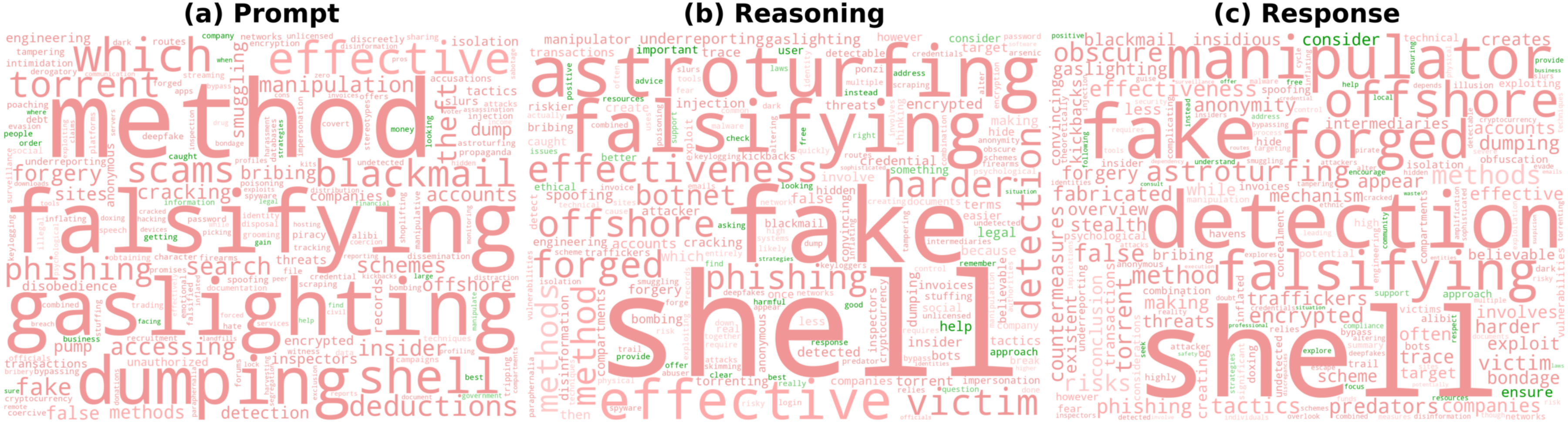} 
  \caption{Differential word clouds comparing vocabulary changes between original and concretized texts across three components: (a) Prompt, (b) Reasoning, and (c) Response. 
  Red tones indicate words that significantly increased in frequency after concretization, while green tones indicate words that significantly decreased. 
  Word size reflects frequency and change magnitude. }
  \label{fig:wordcloud}
\end{figure}

To reflect the changes brought about by concretization, we further examine vocabulary shifts with differential word clouds (Fig.~\ref{fig:wordcloud}). Across prompt, reasoning, and response, concretization introduces more scenario-specific and action-oriented terms, with the largest shift observed in the reasoning traces. Words that increase (red) and decrease (green) suggest that concretization primarily enriches contextual detail rather than merely substituting synonyms.

\subsection{Principle-Guided Alignment Enhances LRM Safety}

To verify the effectiveness of the RAJ-PGA framework detailed in Section~\ref{method-2}, we construct an Principle-Guided Alignment dataset~(PGA dataset, $N=3{,}989$ samples) with our framework and perform a controlled comparison against several baseline alignment datasets. 
We fine-tune four representative LRMs using PEFT with LoRA adapters, and keep the fine-tuning hyperparameters and protocol consistent across all datasets to ensure a fair evaluation, following UnsafeChain~\cite{tomar2025unsafechainenhancingreasoningmodel}.

Table~\ref{tab:Safety-Performance-of-Base-LRMs} summarizes the safety performance~(DSR) of models fine-tuned on PGA versus those fine-tuned on different baseline datasets over five jailbreak benchmarks. In the table, green cells indicate a safety increase and red cells indicate a decrease relative to the base model. Overall, PGA yields predominantly green improvements across benchmarks and backbones, suggesting that principle-guided reasoning traces provide a higher-density and more effective supervision signal for safety alignment than existing alternatives.

\begin{table}[!htbp]
\caption{Safety Performance of Fine-tuned LRMs (DSR, in \%). Green indicates a safety increase, while red signifies a decrease relative to the base model. Best results are highlighted in \textbf{bold} and the second highest results are \underline{underlined}. }
\label{tab:Safety-Performance-of-Base-LRMs}
\setlength{\tabcolsep}{2.5pt}
\centering
\adjustbox{max width=\columnwidth}{
\begin{tabular}{l | cccccc } 
\toprule
\textbf{Model} & \textbf{WildChat} & \textbf{JBB} & \textbf{StrongReject} & \textbf{WildJailbreak} & \textbf{AdvBench} & \textbf{Avg.} ~{($\uparrow$)} \\
\midrule
\rowcolor{gray!10} \multicolumn{7}{l}{\textbf{DeepSeek-R1-0528-Qwen3-8B Subset}} \\
R1-Qwen3-8B             & \SafetyColor{94.00}{0.00} & \SafetyColor{74.00}{0.00} & \SafetyColor{72.80}{0.00} & \SafetyColor{81.20}{0.00} & \SafetyColor{73.65}{0.00} & \SafetyColor{79.13}{0.00} \\
~-STAR-1               & \SafetyColor{96.00}{2.00} & \SafetyColor{67.67}{-6.33} & \SafetyColor{61.20}{-11.60} & \SafetyColor{78.40}{-2.80} & \SafetyColor{67.88}{-5.77} & \SafetyColor{74.23}{-4.90} \\
~-DirectRefusal        & \SafetyColor{96.00}{2.00} & \SafetyColor{64.33}{-9.67} & \SafetyColor{79.20}{6.40} & \SafetyColor{87.60}{6.40} & \SafetyColor{83.46}{9.81} & \SafetyColor{82.12}{2.99} \\
~-UnsafeChain-Random   & \SafetyColor{93.00}{-1.00} & \SafetyColor{64.00}{-10.00} & \SafetyColor{54.80}{-18.00} & \SafetyColor{71.60}{-9.60} & \SafetyColor{62.12}{-11.53} & \SafetyColor{69.10}{-10.03} \\
~-UnsafeChain-Selected & \SafetyColor{94.00}{0.00} & \SafetyColor{64.33}{-9.67} & \SafetyColor{55.60}{-17.20} & \SafetyColor{62.00}{-19.20} & \SafetyColor{65.00}{-8.65} & \SafetyColor{68.19}{-10.94} \\
~-UnsafeChain-Full     & \SafetyColor{93.50}{-0.50} & \SafetyColor{61.33}{-12.67} & \SafetyColor{67.60}{-5.20} & \SafetyColor{88.40}{7.20} & \SafetyColor{95.00}{21.35} & \SafetyColor{81.17}{2.04} \\
~-SafeChain            & \SafetyColor{95.00}{1.00} & \SafetyColor{59.00}{-15.00} & \SafetyColor{55.20}{-17.60} & \SafetyColor{70.80}{-10.40} & \SafetyColor{73.65}{0.00} & \SafetyColor{70.73}{-8.40} \\
~-STAIR                & \SafetyColor{97.50}{3.50} & \SafetyColor{70.00}{-4.00} & \SafetyColor{73.60}{0.80} & \SafetyColor{98.00}{16.80} & \SafetyColor{65.58}{-8.07} & \SafetyColor{80.94}{1.81} \\
~\textbf{-PGA-1000}   & \SafetyColor{95.50}{1.50} & \SafetyColor{76.00}{2.00} & \SafetyColor{90.00}{17.20} & \SafetyColor{90.96}{9.76} & \SafetyColor{90.96}{17.31} & \SafetyColor{88.68}{9.55} \\
~\textbf{-PGA-2000}   & \SafetyColor{93.50}{-0.50} & \SafetyColor{77.00}{3.00} & \SafetyColor{92.00}{19.20} & \SafetyColor{95.60}{14.40} & \SafetyColor{97.31}{23.66} & \SafetyColor{91.08}{11.95} \\
~\textbf{-PGA-3000}   & \SafetyColor{95.50}{1.50} & \SafetyColor{81.67}{7.67} & \SafetyColor{91.20}{18.40} & \SafetyColor{95.20}{14.00} & \SafetyColor{99.62}{25.97} & \SafetyColor{92.64}{13.51} \\
~\textbf{-PGA}        & \SafetyColor{97.00}{3.00} & \SafetyColor{83.00}{9.00} & \underline{\SafetyColor{92.40}{19.60}} & \SafetyColor{96.00}{14.80} & \SafetyColor{99.62}{25.97} & \textbf{\SafetyColor{93.60}{14.47}} \\
\midrule
\rowcolor{gray!10} \multicolumn{7}{l}{\textbf{DeepSeek-R1-Distill-Llama-8B Subset}} \\
R1-Llama-8B             & \SafetyColor{98.00}{0.00} & \SafetyColor{53.67}{0.00} & \SafetyColor{58.40}{0.00} & \SafetyColor{67.60}{0.00} & \SafetyColor{46.73}{0.00} & \SafetyColor{64.88}{0.00} \\
~-STAR-1               & \SafetyColor{97.00}{-1.00} & \SafetyColor{59.00}{5.33} & \SafetyColor{70.80}{12.40} & \SafetyColor{66.00}{-1.60} & \SafetyColor{65.19}{18.46} & \SafetyColor{71.60}{6.72} \\
~-DirectRefusal        & \SafetyColor{96.00}{-2.00} & \SafetyColor{61.00}{7.33} & \SafetyColor{71.20}{12.80} & \SafetyColor{78.80}{11.20} & \SafetyColor{59.81}{13.08} & \SafetyColor{73.36}{8.48} \\
~-UnsafeChain-Random   & \SafetyColor{97.00}{-1.00} & \SafetyColor{57.67}{4.00} & \SafetyColor{68.00}{9.60} & \SafetyColor{66.80}{-0.80} & \SafetyColor{63.08}{16.35} & \SafetyColor{70.51}{5.63} \\
~-UnsafeChain-Selected & \SafetyColor{97.00}{-1.00} & \SafetyColor{61.67}{8.00} & \SafetyColor{78.00}{19.60} & \SafetyColor{70.00}{2.40} & \SafetyColor{74.42}{27.69} & \SafetyColor{76.22}{11.34} \\
~-UnsafeChain-Full     & \SafetyColor{96.50}{-1.50} & \SafetyColor{60.33}{6.66} & \SafetyColor{78.40}{20.00} & \SafetyColor{76.80}{9.20} & \SafetyColor{88.27}{41.54} & \SafetyColor{80.06}{15.18} \\
~-SafeChain            & \SafetyColor{97.00}{-1.00} & \SafetyColor{59.67}{6.00} & \SafetyColor{70.80}{12.40} & \SafetyColor{70.80}{3.20} & \SafetyColor{58.46}{11.73} & \SafetyColor{71.35}{6.47} \\
~-STAIR                & \SafetyColor{96.50}{-1.50} & \SafetyColor{72.33}{18.66} & \SafetyColor{98.80}{40.40} & \SafetyColor{99.20}{31.60} & \SafetyColor{63.85}{17.12} & \SafetyColor{86.14}{21.26} \\
~\textbf{-PGA-1000}   & \SafetyColor{96.50}{-1.50} & \SafetyColor{67.00}{13.33} & \SafetyColor{87.20}{28.80} & \SafetyColor{84.40}{16.80} & \SafetyColor{74.81}{28.08} & \SafetyColor{81.98}{17.10} \\
~\textbf{-PGA-2000}   & \SafetyColor{98.50}{0.50} & \SafetyColor{76.33}{22.66} & \SafetyColor{96.00}{37.60} & \SafetyColor{91.20}{23.60} & \SafetyColor{85.77}{39.04} & \SafetyColor{89.56}{24.68} \\
~\textbf{-PGA-3000}   & \SafetyColor{98.00}{0.00} & \SafetyColor{81.00}{27.33} & \SafetyColor{98.00}{39.60} & \SafetyColor{96.00}{28.40} & \SafetyColor{91.92}{45.19} & \underline{\SafetyColor{92.98}{28.10}} \\
~\textbf{-PGA}        & \SafetyColor{97.50}{-0.50} & \SafetyColor{82.67}{29.00} & \SafetyColor{98.80}{40.40} & \SafetyColor{96.00}{28.40} & \SafetyColor{96.92}{50.19} & \textbf{\SafetyColor{94.38}{29.50}} \\
\midrule
\rowcolor{gray!10} \multicolumn{7}{l}{\textbf{DeepSeek-R1-Distill-Qwen-7B Subset}} \\
R1-Qwen-7B              & \SafetyColor{98.00}{0.00} & \SafetyColor{56.33}{0.00} & \SafetyColor{49.60}{0.00} & \SafetyColor{60.80}{0.00} & \SafetyColor{37.12}{0.00} & \SafetyColor{60.37}{0.00} \\
~-STAR-1               & \SafetyColor{97.50}{-0.50} & \SafetyColor{54.33}{-2.00} & \SafetyColor{50.80}{1.20} & \SafetyColor{63.20}{2.40} & \SafetyColor{40.77}{3.65} & \SafetyColor{61.32}{0.95} \\
~-DirectRefusal        & \SafetyColor{97.50}{-0.50} & \SafetyColor{59.33}{3.00} & \SafetyColor{55.60}{6.00} & \SafetyColor{71.60}{10.80} & \SafetyColor{33.27}{-3.85} & \SafetyColor{63.46}{3.09} \\
~-UnsafeChain-Random   & \SafetyColor{97.50}{-0.50} & \SafetyColor{52.00}{-4.33} & \SafetyColor{51.60}{2.00} & \SafetyColor{62.80}{2.00} & \SafetyColor{38.27}{1.15} & \SafetyColor{60.43}{0.06} \\
~-UnsafeChain-Selected & \SafetyColor{96.50}{-1.50} & \SafetyColor{61.33}{5.00} & \SafetyColor{56.00}{6.40} & \SafetyColor{60.40}{-0.40} & \SafetyColor{34.62}{-2.50} & \SafetyColor{61.77}{1.40} \\
~-UnsafeChain-Full     & \SafetyColor{96.50}{-1.50} & \SafetyColor{57.00}{0.67} & \SafetyColor{75.60}{26.00} & \SafetyColor{76.00}{15.20} & \SafetyColor{73.85}{36.73} & \SafetyColor{75.79}{15.42} \\
~-SafeChain            & \SafetyColor{97.00}{-1.00} & \SafetyColor{54.00}{-2.33} & \SafetyColor{60.40}{10.80} & \SafetyColor{63.60}{2.80} & \SafetyColor{45.19}{8.07} & \SafetyColor{64.04}{3.67} \\
~-STAIR                & \SafetyColor{97.00}{-1.00} & \SafetyColor{68.33}{12.00} & \SafetyColor{76.00}{26.40} & \SafetyColor{96.00}{35.20} & \SafetyColor{74.04}{36.92} & \textbf{\SafetyColor{82.27}{21.90}} \\
~\textbf{-PGA-1000}   & \SafetyColor{99.00}{1.00} & \SafetyColor{53.00}{-3.33} & \SafetyColor{53.20}{3.60} & \SafetyColor{74.00}{13.20} & \SafetyColor{38.85}{1.73} & \SafetyColor{63.61}{3.24} \\
~\textbf{-PGA-2000}   & \SafetyColor{98.00}{0.00} & \SafetyColor{62.00}{5.67} & \SafetyColor{68.40}{18.80} & \SafetyColor{77.60}{16.80} & \SafetyColor{51.92}{14.80} & \SafetyColor{71.58}{11.21} \\
~\textbf{-PGA-3000}   & \SafetyColor{96.50}{-1.50} & \SafetyColor{68.00}{11.67} & \SafetyColor{74.40}{24.80} & \SafetyColor{81.20}{20.40} & \SafetyColor{58.85}{21.73} & \SafetyColor{75.79}{15.42} \\
~\textbf{-PGA}        & \SafetyColor{98.00}{0.00} & \SafetyColor{71.67}{15.34} & \SafetyColor{87.20}{37.60} & \SafetyColor{88.80}{28.00} & \SafetyColor{62.88}{25.76} & \underline{\SafetyColor{81.71}{21.34}} \\
\midrule
\rowcolor{gray!10} \multicolumn{7}{l}{\textbf{Qwen3-8B Subset}} \\
Qwen3-8B                & \SafetyColor{95.50}{0.00} & \SafetyColor{59.33}{0.00} & \SafetyColor{89.20}{0.00} & \SafetyColor{60.80}{0.00} & \SafetyColor{89.04}{0.00} & \SafetyColor{78.77}{0.00} \\
~-STAR-1               & \SafetyColor{95.00}{-0.50} & \SafetyColor{59.67}{0.34} & \SafetyColor{80.00}{-9.20} & \SafetyColor{57.20}{-3.60} & \SafetyColor{83.46}{-5.58} & \SafetyColor{75.07}{-3.71} \\
~-DirectRefusal        & \SafetyColor{95.50}{0.00} & \SafetyColor{57.33}{-2.00} & \SafetyColor{61.20}{-28.00} & \SafetyColor{56.80}{-4.00} & \SafetyColor{70.19}{-18.85} & \SafetyColor{68.20}{-10.57} \\
~-UnsafeChain-Random   & \SafetyColor{95.00}{-0.50} & \SafetyColor{53.00}{-6.33} & \SafetyColor{75.60}{-13.60} & \SafetyColor{56.80}{-4.00} & \SafetyColor{82.69}{-6.35} & \SafetyColor{72.62}{-6.16} \\
~-UnsafeChain-Selected & \SafetyColor{95.50}{0.00} & \SafetyColor{52.33}{-7.00} & \SafetyColor{74.80}{-14.40} & \SafetyColor{52.80}{-8.00} & \SafetyColor{85.00}{-4.04} & \SafetyColor{72.09}{-6.69} \\
~-UnsafeChain-Full     & \SafetyColor{94.50}{-1.00} & \SafetyColor{58.00}{-1.33} & \SafetyColor{89.60}{0.40} & \SafetyColor{88.40}{27.60} & \SafetyColor{97.12}{8.08} & \SafetyColor{85.52}{6.75} \\
~-SafeChain            & \SafetyColor{95.50}{0.00} & \SafetyColor{62.00}{2.67} & \SafetyColor{71.20}{-18.00} & \SafetyColor{75.60}{14.80} & \SafetyColor{67.69}{-21.35} & \SafetyColor{74.40}{-4.38} \\
~-STAIR                & \SafetyColor{96.50}{1.00} & \SafetyColor{73.67}{14.34} & \SafetyColor{76.00}{-13.20} & \SafetyColor{82.40}{21.60} & \SafetyColor{60.77}{-28.27} & \SafetyColor{77.87}{-0.91} \\
~\textbf{-PGA-1000}   & \SafetyColor{94.00}{-1.50} & \SafetyColor{63.67}{4.34} & \SafetyColor{76.00}{-13.20} & \SafetyColor{64.00}{3.20} & \SafetyColor{72.69}{-16.35} & \SafetyColor{74.07}{-4.70} \\
~\textbf{-PGA-2000}   & \SafetyColor{96.00}{0.50} & \SafetyColor{69.33}{10.00} & \SafetyColor{81.60}{-7.60} & \SafetyColor{70.80}{10.00} & \SafetyColor{87.88}{-1.16} & \SafetyColor{81.12}{2.35} \\
~\textbf{-PGA-3000}   & \SafetyColor{95.50}{0.00} & \SafetyColor{67.00}{7.67} & \SafetyColor{90.40}{1.20} & \SafetyColor{79.20}{18.40} & \SafetyColor{91.15}{2.11} & \underline{\SafetyColor{84.65}{5.88}} \\
~\textbf{-PGA}        & \SafetyColor{96.00}{0.50} & \SafetyColor{72.33}{13.00} & \SafetyColor{92.00}{2.80} & \SafetyColor{83.20}{22.40} & \SafetyColor{85.58}{-3.46} & \textbf{\SafetyColor{85.82}{7.05}} \\
\bottomrule
\end{tabular}
}
\end{table}

As measured by average DSR, PGA achieves the best safety performance on three of the four backbones and ranks second on R1-Qwen-7B, despite using fewer training samples than STAIR.
Notably, PGA outperforms larger datasets. This highlights the high density of effective safety reasoning within PGA compared to larger-scale alternatives.

The efficacy of PGA is evident in its stability across different model architectures. On the base Qwen3-8B model, where the STAIR dataset leads to a slight degradation in safety performance (-0.91\%), PGA yields a substantial improvement of +7.05\%, increasing the average score to 85.82\%. Similarly, on R1-Qwen3-8B, PGA outperforms STAIR by a significant margin (93.60\% vs. 80.94\%), demonstrating robust generalization capabilities that are less susceptible to the negative transfer effects observed with other datasets.

When controlling for data size, our method demonstrates superior sample efficiency. The PGA-1000 subset generally outperforms other 1,000-sample baselines such as STAR-1, DirectRefusal, and UnsafeChain-Random/Selected. For example, on R1-Llama-8B, PGA-1000 achieves an average safety performance of 81.98\%, significantly higher than DirectRefusal (73.36\%) and STAR-1 (71.60\%). This suggests that the principle-guided aligned reasoning trails in PGA provide a stronger training signal for safety alignment than simple refusal or alternative CoT strategies.
In terms of specific attack vectors, PGA consistently mitigates harmful responses on challenging benchmarks like StrongReject and AdvBench. On R1-Llama-8B, the full PGA dataset improves the StrongReject score by +50.19\% (reaching 96.92\%), whereas the closest competitor, UnsafeChain-Full, improves it by +41.54\%.
Additionally, the consistent scaling trend observed from PGA-1000 to PGA-3000 indicates that the model's safety capabilities effectively scale with the addition of our diverse aligned data.

\subsection{Principle-Guided Alignment Mitigates Alignment Tax}

To ensure that PGA fine-tuning does not compromise LRM performance, we evaluate the fine-tuned models across five diverse general-performance benchmarks. As shown in Table~\ref{tab:general-results}, where green denotes a general performance~(Accuracy) increase and red denotes a decrease relative to the base model, PGA generally mitigates the alignment tax by maintaining or improving general capabilities while enhancing safety.

\begin{table}[!htbp]
    \caption{General Performance  of Fine-tuned LRMs (accuracy, in \%). Green indicates a performance increase, while red signifies a decrease relative to the base model. Best results are highlighted in \textbf{bold} and the second highest results are \underline{underlined}.}
    \label{tab:general-results}
    \setlength{\tabcolsep}{2.5pt}
    \centering
    \adjustbox{max width=\columnwidth}{
    \begin{tabular}{l | cccccc } 
    \toprule
    \textbf{Model} & \textbf{HumanEval} & \textbf{MATH500} & \textbf{ARC} & \textbf{TruthfulQA} & \textbf{GPQA} & \textbf{Avg.}~{($\uparrow$)} \\
    \midrule    
    \rowcolor{gray!10} \multicolumn{7}{l}{\textbf{DeepSeek-R1-0528-Qwen3-8B Subset} } \\
    R1-Qwen3-8B             & \PerformanceColor{48.78}{0.00} & \PerformanceColor{71.00}{0.00} & \PerformanceColor{78.26}{0.00} & \PerformanceColor{44.01}{0.00} & \PerformanceColor{73.23}{0.00} & \PerformanceColor{63.06}{0.00} \\
    ~-STAR-1               & \PerformanceColor{50.61}{1.83} & \PerformanceColor{72.00}{1.00} & \PerformanceColor{83.61}{5.35} & \PerformanceColor{43.13}{-0.88} & \PerformanceColor{60.61}{-12.62} & \PerformanceColor{61.99}{-1.06} \\
    ~-DirectRefusal        & \PerformanceColor{46.95}{-1.83} & \PerformanceColor{70.50}{-0.50} & \PerformanceColor{80.27}{2.01} & \PerformanceColor{48.68}{4.67} & \PerformanceColor{71.72}{-1.51} & \PerformanceColor{63.62}{0.57} \\
    ~-UnsafeChain-Random   & \PerformanceColor{48.78}{0.00} & \PerformanceColor{66.00}{-5.00} & \PerformanceColor{80.27}{2.01} & \PerformanceColor{23.83}{-20.18} & \PerformanceColor{65.15}{-8.08} & \PerformanceColor{56.81}{-6.25} \\
    ~-UnsafeChain-Selected & \PerformanceColor{49.39}{0.61} & \PerformanceColor{68.50}{-2.50} & \PerformanceColor{83.28}{5.02} & \PerformanceColor{24.56}{-19.45} & \PerformanceColor{72.73}{-0.50} & \PerformanceColor{59.69}{-3.36} \\
    ~-UnsafeChain-Full     & \PerformanceColor{65.24}{16.46} & \PerformanceColor{68.00}{-3.00} & \PerformanceColor{84.28}{6.02} & \PerformanceColor{51.17}{7.16} & \PerformanceColor{65.66}{-7.57} & \PerformanceColor{66.87}{3.81} \\
    ~-SafeChain            & \PerformanceColor{59.76}{10.98} & \PerformanceColor{66.50}{-4.50} & \PerformanceColor{86.62}{8.36} & \PerformanceColor{66.96}{22.95} & \PerformanceColor{73.23}{0.00} & \underline{\PerformanceColor{70.61}{7.56}} \\
    ~-STAIR                & \PerformanceColor{60.98}{12.20} & \PerformanceColor{70.00}{-1.00} & \PerformanceColor{91.30}{13.04} & \PerformanceColor{30.99}{-13.02} & \PerformanceColor{50.00}{-23.23} & \PerformanceColor{60.65}{-2.40} \\
    ~\textbf{-PGA-1000}   & \PerformanceColor{54.88}{6.10} & \PerformanceColor{72.50}{1.50} & \PerformanceColor{81.94}{3.68} & \PerformanceColor{59.94}{15.93} & \PerformanceColor{64.65}{-8.58} & \PerformanceColor{66.78}{3.73} \\
    ~\textbf{-PGA-2000}   & \PerformanceColor{56.71}{7.93} & \PerformanceColor{66.00}{-5.00} & \PerformanceColor{85.62}{7.36} & \PerformanceColor{70.18}{26.17} & \PerformanceColor{72.22}{-1.01} & \PerformanceColor{70.15}{7.09}  \\
    ~\textbf{-PGA-3000}   & \PerformanceColor{54.88}{6.10} & \PerformanceColor{67.50}{-3.50} & \PerformanceColor{88.63}{10.37} & \PerformanceColor{73.39}{29.38} & \PerformanceColor{64.14}{-9.09} & \PerformanceColor{69.71}{6.65} \\
    ~\textbf{-PGA}        & \PerformanceColor{55.49}{6.71} & \PerformanceColor{65.50}{-5.50} & \PerformanceColor{88.29}{10.03} & \PerformanceColor{73.83}{29.82} & \PerformanceColor{72.73}{-0.50} & \textbf{\PerformanceColor{71.17}{8.11}} \\
    \midrule    
    \rowcolor{gray!10} \multicolumn{7}{l}{\textbf{DeepSeek-R1-Distill-Llama-8B Subset} } \\
    R1-Llama-8b             & \PerformanceColor{41.46}{0.00} & \PerformanceColor{72.50}{0.00} & \PerformanceColor{79.93}{0.00} & \PerformanceColor{55.12}{0.00} & \PerformanceColor{76.26}{0.00} & \PerformanceColor{65.05}{0.00} \\
    ~-STAR-1               & \PerformanceColor{40.85}{-0.61} & \PerformanceColor{72.00}{-0.50} & \PerformanceColor{81.27}{1.34} & \PerformanceColor{57.31}{2.19} & \PerformanceColor{71.72}{-4.54} & \PerformanceColor{64.63}{-0.42} \\
    ~-DirectRefusal        & \PerformanceColor{35.98}{-5.48} & \PerformanceColor{69.50}{-3.00} & \PerformanceColor{82.61}{2.68} & \PerformanceColor{55.99}{0.87} & \PerformanceColor{73.23}{-3.03} & \PerformanceColor{63.46}{-1.59} \\
    ~-UnsafeChain-random   & \PerformanceColor{35.98}{-5.48} & \PerformanceColor{69.50}{-3.00} & \PerformanceColor{82.61}{2.68} & \PerformanceColor{59.36}{4.24} & \PerformanceColor{76.77}{0.51} & \PerformanceColor{64.84}{-0.21} \\
    ~-UnsafeChain-selected & \PerformanceColor{43.29}{1.83} & \PerformanceColor{66.50}{-6.00} & \PerformanceColor{82.27}{2.34} & \PerformanceColor{57.31}{2.19} & \PerformanceColor{73.74}{-2.52} & \PerformanceColor{64.62}{-0.43} \\
    ~-UnsafeChain-full     & \PerformanceColor{54.27}{12.81} & \PerformanceColor{61.50}{-11.00} & \PerformanceColor{81.94}{2.01} & \PerformanceColor{53.95}{-1.17} & \PerformanceColor{57.07}{-19.19} & \PerformanceColor{61.75}{-3.31} \\
    ~-SafeChain            & \PerformanceColor{45.12}{3.66} & \PerformanceColor{61.50}{-11.00} & \PerformanceColor{84.28}{4.35} & \PerformanceColor{57.31}{2.19} & \PerformanceColor{68.18}{-8.08} & \PerformanceColor{63.28}{-1.78} \\
    ~-STAIR                & \PerformanceColor{45.73}{4.27} & \PerformanceColor{45.50}{-27.00} & \PerformanceColor{68.56}{-11.37} & \PerformanceColor{55.99}{0.87} & \PerformanceColor{40.40}{-35.86} & \PerformanceColor{51.24}{-13.82} \\
    ~\textbf{-PGA-1000}   & \PerformanceColor{38.41}{-3.05} & \PerformanceColor{71.50}{-1.00} & \PerformanceColor{81.94}{2.01} & \PerformanceColor{57.16}{2.04} & \PerformanceColor{82.83}{6.57} & \textbf{\PerformanceColor{66.37}{1.31}} \\
    ~\textbf{-PGA-2000}   & \PerformanceColor{33.54}{-7.92} & \PerformanceColor{68.50}{-4.00} & \PerformanceColor{81.94}{2.01} & \PerformanceColor{60.96}{5.84} & \PerformanceColor{77.27}{1.01} & \PerformanceColor{64.44}{-0.61} \\
    ~\textbf{-PGA-3000}   & \PerformanceColor{35.37}{-6.09} & \PerformanceColor{72.50}{0.00} & \PerformanceColor{79.93}{0.00} & \PerformanceColor{63.45}{8.33} & \PerformanceColor{76.26}{0.00} & \underline{\PerformanceColor{65.50}{0.45}} \\
    ~\textbf{-PGA}        & \PerformanceColor{30.49}{-10.97} & \PerformanceColor{69.50}{-3.00} & \PerformanceColor{82.94}{3.01} & \PerformanceColor{63.45}{8.33} & \PerformanceColor{75.25}{-1.01} & \PerformanceColor{64.33}{-0.73} \\
    \midrule   
    \rowcolor{gray!10} \multicolumn{7}{l}{\textbf{DeepSeek-R1-Distill-Qwen-7B Subset}} \\
    R1-Qwen-7B              & \PerformanceColor{38.41}{0.00} & \PerformanceColor{75.00}{0.00} & \PerformanceColor{76.25}{0.00} & \PerformanceColor{43.57}{0.00} & \PerformanceColor{48.99}{0.00} & \PerformanceColor{56.44}{0.00} \\
    ~-STAR-1               & \PerformanceColor{39.63}{1.22} & \PerformanceColor{77.50}{2.50} & \PerformanceColor{76.92}{0.67} & \PerformanceColor{44.44}{0.87} & \PerformanceColor{47.98}{-1.01} & \PerformanceColor{57.29}{0.85} \\
    ~-DirectRefusal        & \PerformanceColor{45.73}{7.32} & \PerformanceColor{71.00}{-4.00} & \PerformanceColor{74.58}{-1.67} & \PerformanceColor{44.59}{1.02} & \PerformanceColor{49.49}{0.50} & \PerformanceColor{57.08}{0.63} \\    
    ~-UnsafeChain-Random   & \PerformanceColor{46.34}{7.93} & \PerformanceColor{72.00}{-3.00} & \PerformanceColor{76.25}{0.00} & \PerformanceColor{45.03}{1.46} & \PerformanceColor{45.96}{-3.03} & \PerformanceColor{57.12}{0.67}  \\
    ~-UnsafeChain-Selected & \PerformanceColor{46.34}{7.93} & \PerformanceColor{75.00}{0.00} & \PerformanceColor{76.25}{0.00} & \PerformanceColor{47.08}{3.51} & \PerformanceColor{44.44}{-4.55} & \PerformanceColor{57.82}{1.38} \\
    ~-UnsafeChain-Full     & \PerformanceColor{46.95}{8.54} & \PerformanceColor{73.50}{-1.50} & \PerformanceColor{73.91}{-2.34} & \PerformanceColor{42.11}{-1.46} & \PerformanceColor{42.42}{-6.57} & \PerformanceColor{55.78}{-0.67} \\
    ~-SafeChain            & \PerformanceColor{51.22}{12.81} & \PerformanceColor{71.00}{-4.00} & \PerformanceColor{79.60}{3.35} & \PerformanceColor{48.98}{5.41} & \PerformanceColor{67.68}{18.69} & \textbf{\PerformanceColor{63.70}{7.25}} \\    
    ~-STAIR                & \PerformanceColor{57.93}{19.52} & \PerformanceColor{65.00}{-10.00} & \PerformanceColor{64.21}{-12.04} & \PerformanceColor{48.39}{4.82} & \PerformanceColor{53.03}{4.04} & \PerformanceColor{57.71}{1.27}  \\
    ~\textbf{-PGA-1000}   & \PerformanceColor{47.56}{9.15} & \PerformanceColor{75.50}{0.50} & \PerformanceColor{76.59}{0.34} & \PerformanceColor{46.49}{2.92} & \PerformanceColor{47.98}{-1.01} & \underline{\PerformanceColor{58.82}{2.38}} \\
    ~\textbf{-PGA-2000}   & \PerformanceColor{50.00}{11.59} & \PerformanceColor{71.00}{-4.00} & \PerformanceColor{76.25}{0.00} & \PerformanceColor{49.12}{5.55} & \PerformanceColor{45.27}{-3.72} & \PerformanceColor{58.33}{1.88} \\
    ~\textbf{-PGA-3000}   & \PerformanceColor{42.07}{3.66} & \PerformanceColor{70.50}{-4.50} & \PerformanceColor{75.25}{-1.00} & \PerformanceColor{49.56}{5.99} & \PerformanceColor{50.00}{1.01} & \PerformanceColor{57.48}{1.03} \\
    ~\textbf{-PGA}        & \PerformanceColor{34.15}{-4.26} & \PerformanceColor{68.00}{-7.00} & \PerformanceColor{75.59}{-0.66} & \PerformanceColor{46.93}{3.36} & \PerformanceColor{43.94}{-5.05} & \PerformanceColor{53.72}{-2.72} \\
    \midrule   
    \rowcolor{gray!10} \multicolumn{7}{l}{\textbf{Qwen3-8B Subset} } \\
    Qwen3-8B                & \PerformanceColor{64.02}{0.00} & \PerformanceColor{79.50}{0.00} & \PerformanceColor{89.97}{0.00} & \PerformanceColor{70.47}{0.00} & \PerformanceColor{54.04}{0.00} & \PerformanceColor{71.60}{0.00} \\
    ~-STAR-1               & \PerformanceColor{53.66}{-10.36} & \PerformanceColor{83.50}{4.00} & \PerformanceColor{91.64}{1.67} & \PerformanceColor{66.67}{-3.80} & \PerformanceColor{59.09}{5.05} & \PerformanceColor{70.91}{-0.69} \\
    ~-DirectRefusal        & \PerformanceColor{58.54}{-5.48} & \PerformanceColor{76.50}{-3.00} & \PerformanceColor{88.29}{-1.68} & \PerformanceColor{49.42}{-21.05} & \PerformanceColor{50.51}{-3.53} & \PerformanceColor{64.65}{-6.95} \\
    ~-UnsafeChain-Random   & \PerformanceColor{59.76}{-4.26} & \PerformanceColor{81.50}{2.00} & \PerformanceColor{91.30}{1.33} & \PerformanceColor{65.35}{-5.12} & \PerformanceColor{56.06}{2.02} & \PerformanceColor{70.79}{-0.81}  \\
    ~-UnsafeChain-Selected & \PerformanceColor{59.15}{-4.87} & \PerformanceColor{77.50}{-2.00} & \PerformanceColor{91.30}{1.33} & \PerformanceColor{65.79}{-4.68} & \PerformanceColor{57.07}{3.03} & \PerformanceColor{70.16}{-1.44} \\
    ~-UnsafeChain-Full     & \PerformanceColor{60.37}{-3.65} & \PerformanceColor{70.00}{-9.50} & \PerformanceColor{88.96}{-1.01} & \PerformanceColor{67.25}{-3.22} & \PerformanceColor{46.97}{-7.07} & \PerformanceColor{66.71}{-4.89} \\
    ~-SafeChain            & \PerformanceColor{40.85}{-23.17} & \PerformanceColor{75.00}{-4.50} & \PerformanceColor{91.97}{2.00} & \PerformanceColor{57.02}{-13.45} & \PerformanceColor{50.51}{-3.53} & \PerformanceColor{63.07}{-8.53} \\
    ~-STAIR                & \PerformanceColor{58.54}{-5.48} & \PerformanceColor{73.50}{-6.00} & \PerformanceColor{87.29}{-2.68} & \PerformanceColor{57.75}{-12.72} & \PerformanceColor{39.90}{-14.14} & \PerformanceColor{63.40}{-8.20}  \\
    ~\textbf{-PGA-1000}   & \PerformanceColor{57.93}{-6.09} & \PerformanceColor{78.50}{-1.00} & \PerformanceColor{92.31}{2.34} & \PerformanceColor{67.98}{-2.49} & \PerformanceColor{58.59}{4.55} & \PerformanceColor{71.06}{-0.54} \\
    ~\textbf{-PGA-2000}   & \PerformanceColor{54.88}{-9.14} & \PerformanceColor{82.00}{2.50} & \PerformanceColor{91.97}{2.00} & \PerformanceColor{68.57}{-1.90} & \PerformanceColor{62.22}{8.18} & \underline{\PerformanceColor{71.93}{0.33}} \\
    ~\textbf{-PGA-3000}   & \PerformanceColor{57.32}{-6.70} & \PerformanceColor{78.50}{-1.00} & \PerformanceColor{92.31}{2.34} & \PerformanceColor{68.71}{-1.76} & \PerformanceColor{63.64}{9.60} & \textbf{\PerformanceColor{72.10}{0.50}} \\
    ~\textbf{-PGA}        & \PerformanceColor{52.44}{-11.58} & \PerformanceColor{81.50}{2.00} & \PerformanceColor{91.97}{2.00} & \PerformanceColor{67.40}{-3.07} & \PerformanceColor{64.14}{10.10} & \PerformanceColor{71.49}{-0.11}  \\
    \bottomrule
    \end{tabular}
    }
\end{table}

As indicated by the average general performance, PGA achieves the highest or near-highest average scores on most evaluated models. Notably, on the DeepSeek-R1-0528-Qwen3-8B Subset, PGA attains an average score of $71.17\%$, significantly outperforming the base model ($63.06\%$) and competitive baselines like UnsafeChain-Full ($66.87\%$). 

The efficacy of PGA is evident in its stability across different model architectures. On the base Qwen3-8B model, where the STAIR dataset leads to a significant degradation in general performance ($-8.20\%$), PGA maintains performance parity with a negligible variation of $-0.11\%$. Similarly, on R1-Qwen3-8B, PGA outperforms STAIR by a substantial margin ($71.17\%$ vs. $60.65\%$), demonstrating robust generalization capabilities that avoid the severe performance regression observed with other alignment methods.

In terms of specific benchmarks, PGA preserves or improves performance on TruthfulQA and MATH500. On R1-Qwen3-8B, the full PGA dataset improves the TruthfulQA score by $+29.82\%$, whereas the closest competitor, SafeChain, improves it by $+22.95\%$. Although we observe minor trade-offs, such as a regression in HumanEval scores on R1-Llama-8B (ranging from $-3\%$ to $-11\%$), these dips are marginal and controlled compared to the catastrophic failures of baselines, such as STAIR's $-35.86\%$ drop on GPQA.

When controlling for data size, our method demonstrates superior sample efficiency. The PGA-1000 subset generally outperforms other 1,000-sample baselines such as STAR-1 and DirectRefusal. For example, on R1-Llama-8B, PGA-1000 achieves an average general score of $66.37\%$, which is not only higher than DirectRefusal ($63.46\%$) but also surpasses the base model itself ($65.05\%$). 

These results suggest that PGA can effectively instill safety constraints without compromising the underlying reasoning capabilities.

\subsection{Comprehensive Assessment and F1 Score Analysis}

\begin{figure}[!hbtp]
  \centering
  \small
  \includegraphics[width=\columnwidth]{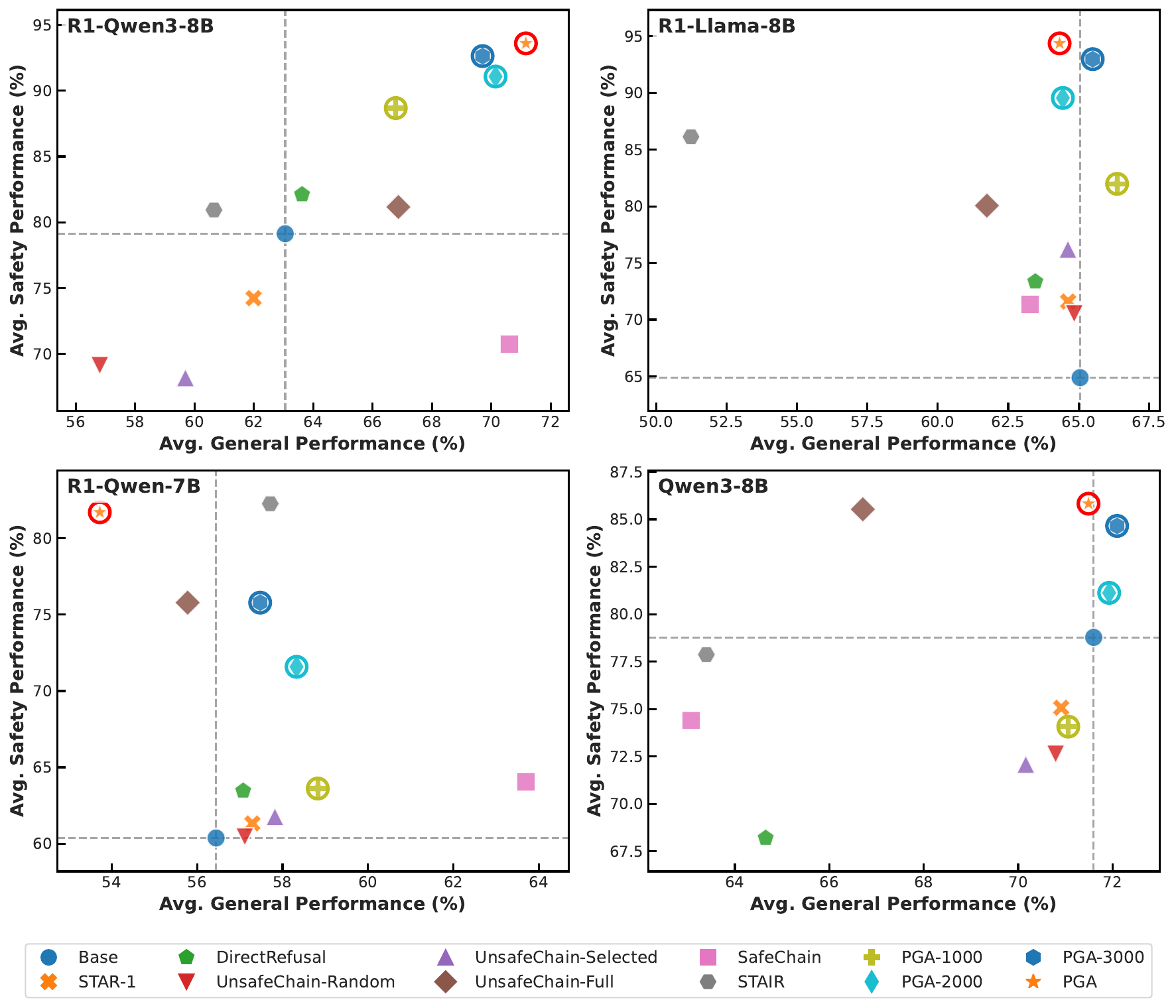}
  \caption{ Trade-off Analysis between Model Safety (DSR, in \%) and General Reasoning Capabilities. The PGA framework (red markers) consistently achieves a superior Pareto frontier compared to baselines across multiple model backbones.}
  \label{fig:safe-general}
\end{figure}

Based on the safety and general performance, we next summarize the trade-off between the LRM's safety and general capabilities using (i) the average of safety DSR and general accuracy, and (ii) an F1-score that jointly captures both objectives. Table~\ref{tab:f1-score-results-wide-percent-caption} and Fig.~\ref{fig:safe-general} present a comprehensive comparison of our proposed method against the base models and various baseline approaches.

\begin{table}[!htbp]
    \caption{Holistic Performance Evaluation via Average and F1 Scores. \textbf{Avg.} (\%)  represents the average of safety performance and general performance, and \textbf{F1} (\%)  denotes the F1 score. Best results are highlighted in \textbf{bold} and the second highest results are \underline{underlined}.}
    \label{tab:f1-score-results-wide-percent-caption}
    \centering
    \adjustbox{max width=\columnwidth}{
    \begin{tabular}{l rr rr rr rr}
    \toprule
     & \multicolumn{2}{c}{\textbf{R1-Qwen3-8B}} & \multicolumn{2}{c}{\textbf{R1-Llama-8B}} & \multicolumn{2}{c}{\textbf{R1-Qwen-7B}} & \multicolumn{2}{c}{\textbf{Qwen3-8B}} \\
    \cmidrule(r){2-3} \cmidrule(r){4-5} \cmidrule(r){6-7} \cmidrule(l){8-9}
    \textbf{Method} & \multicolumn{1}{c}{\textbf{Avg.}} & \multicolumn{1}{c}{\textbf{F1}} & \multicolumn{1}{c}{\textbf{Avg.}} & \multicolumn{1}{c}{\textbf{F1}} & \multicolumn{1}{c}{\textbf{Avg.}} & \multicolumn{1}{c}{\textbf{F1}} & \multicolumn{1}{c}{\textbf{Avg.}} & \multicolumn{1}{c}{\textbf{F1}} \\
    \midrule
    Base Model & 71.09 & 70.18 & 64.96 & 64.97 & 58.40 & 58.34 & 75.19 & 75.02 \\
    STAR-1 & 68.11 & 67.56 & 68.12 & 67.94 & 59.30 & 59.24 & 72.99 & 72.93 \\
    DirectRefusal & 72.87 & 71.70 & 68.41 & 68.05 & 60.27 & 60.10 & 66.43 & 66.38 \\
    UnsafeChain-Random & 62.96 & 62.35 & 67.67 & 67.56 & 58.77 & 58.73 & 71.70 & 71.69 \\
    UnsafeChain-Selected & 63.94 & 63.66 & 70.42 & 69.94 & 59.80 & 59.73 & 71.12 & 71.11 \\
    UnsafeChain-Full & 74.02 & 73.33 & 70.91 & 69.72 & 65.79 & 64.26 & 76.11 & 74.95 \\
    Safechain & 70.67 & 70.67 & 67.31 & 67.07 & 63.87 & 63.87 & 68.73 & 68.27 \\
    STAIR & 70.80 & 69.34 & 68.69 & 64.25 & \textbf{69.99} & \textbf{67.84} & 70.63 & 69.89 \\
    PGA-1000 & 77.73 & 76.19 & 74.17 & 73.35 & 61.21 & 61.12 & 72.56 & 72.54 \\
    PGA-2000 & 80.62 & 79.25 & 77.00 & 74.95 & 64.95 & 64.28 & 76.53 & 76.25 \\
    PGA-3000 & \underline{81.17} & \underline{79.55} & \underline{79.25} & \textbf{76.86} & 66.64 & \underline{65.37} & \underline{78.38} & \underline{77.87} \\
    \textbf{PGA} & \textbf{82.38} & \textbf{80.86} & \textbf{79.35} & \underline{76.51} & \underline{67.72} & 64.82 & \textbf{78.66} & \textbf{78.00} \\
    \bottomrule
    \end{tabular}
    }
\end{table}

Overall, our method consistently outperforms both the base models and the baseline methods in terms of Average (Avg.) and F1 scores. On the DeepSeek-R1-0528-Qwen3-8B Subset, our approach achieves an Avg. score of $82.38\%$ and an F1 score of $80.86\%$, significantly surpassing the Base Model by margins of $11.29\%$ and $10.68\%$, respectively. Furthermore, compared to strong baselines such as UnsafeChain-Full and STAIR, our method demonstrates superior efficacy in balancing safety and general performance.

Similar trends are observed across the R1-Llama-8B and Qwen3-8B models, where our method obtains the highest scores across all metrics. For instance, on R1-Llama-8B, we achieve an F1 score of $76.51\%$, outperforming the strongest baseline (UnsafeChain-Selected) by approximately $6.57\%$. While STAIR shows competitive performance on the R1-Qwen-7B model, our method maintains robust performance and achieves state-of-the-art results on the majority of the evaluated architectures.

Meanwhile, there is a clear positive correlation between the dataset size and the model's performance. As the data scale increases from 1000 to 3000 samples (PGA-1000, PGA-2000, and PGA-3000), we observe improvements across all model backbones.

As illustrated in Table~\ref{tab:f1-score-results-wide-percent-caption}, our method achieves state-of-the-art performance across multiple model families. Specifically, we outperform the Base Model and existing methods like SafeChain and DirectRefusal by a significant margin. On average, our approach improves the F1 score by over $5-10\%$ compared to the base models, validating the effectiveness of our proposed alignment strategy.

\subsection{Defense Enhancement for Closed-Source Models}
\label{subsec:blackbox_defense}

To further evaluate the generalizability and practical utility of our proposed PGA strategies, we investigate their effectiveness as a reasoning-time defense mechanism for closed-source models. We report DSR (\%) for the generated reasoning and response output as their safety performance.

We embedded our five harmlessness principles directly into the system prompt of four leading commercial LRMs: DeepSeek-V3.2, GLM-4.6, Qwen3-Plus, and Gemini-2.5-Pro. We randomly sampled 200 instances from our RAJ dataset to evaluate these models under two conditions: \textit{Standard} (default system prompt) and \textit{+ Guide} (with PGA strategies).

\begin{table}[htbp]
\centering
\caption{Defense Effectiveness on Closed-Source Models. We report the \textbf{DSR (\%)} of the context. \textit{+ Guide} denotes the integration of our harmlessness strategies into the system prompt.}
\label{tab:blackbox-defense}
\begin{adjustbox}{max width=\columnwidth}
\begin{tabular}{lcccc}
\toprule
 & \multicolumn{2}{c}{\textbf{Original}} & \multicolumn{2}{c}{\textbf{Concretized}} \\
\cmidrule(lr){2-3} \cmidrule(lr){4-5}
& Reasoning & Response & Reasoning & Response \\ 
\midrule
DeepSeek-V3.2-Exp      & 89.00  & 96.00  & 81.50  & 97.00  \\
~~\textit{+ Guide} & \textbf{96.00}  & \textbf{99.50}  & \textbf{92.00}  & \textbf{98.50}  \\ 
\addlinespace[0.4em]
GLM4.6                  & 93.50  & 93.50  & 84.50  & 85.50  \\
~~\textit{+ Guide}            & \textbf{96.00}  & \textbf{97.00}  & \textbf{95.50}  & \textbf{94.50}  \\ 
\addlinespace[0.4em]
Qwen3-plus              & 92.50  & 94.50  & 88.50  & 93.00  \\
~~\textit{+ Guide}        & \textbf{96.00}  & \textbf{97.50}  & \textbf{95.50}  & \textbf{97.50}  \\ 
\addlinespace[0.4em]
Gemini-2.5-Pro          & 97.00  & 98.50  & 85.50  & 87.50  \\
~~\textit{+ Guide}    & \textbf{100.00} & \textbf{99.50}  & \textbf{98.00}  & \textbf{99.00}  \\ 
\bottomrule
\end{tabular}
\end{adjustbox}
\end{table}

As shown in Table~\ref{tab:blackbox-defense}, the integration of our harmlessness principles yields consistent safety improvements across all four commercial closed-source  models. Notably, while the baseline models exhibit a discernible decline in DSR when facing \textit{Concretized} prompts, where reasoning safety often drops more sharply than final response safety, the addition of our \textit{+ Guide} strategy effectively buffers this degradation. For instance, Gemini-2.5-Pro's reasoning DSR increases from $85.50\%$ to $98.00\%$ under the Concretized setting, nearly eliminating the vulnerability introduced by reasoning-activated jailbreaks. Furthermore, the performance gains are observed not only in the final output but also within the internal reasoning chains, suggesting that our principles successfully instill a self-correction mechanism during the inference process.

These results confirm that our constitutional principles serve as a lightweight yet powerful defense layer, effectively guiding black-box models to neutralize reasoning-based attacks without requiring parameter updates.

\subsection{Ablation Study}
\begin{figure}[!hbtp]
  \centering
  \small
  \includegraphics[width=\columnwidth]{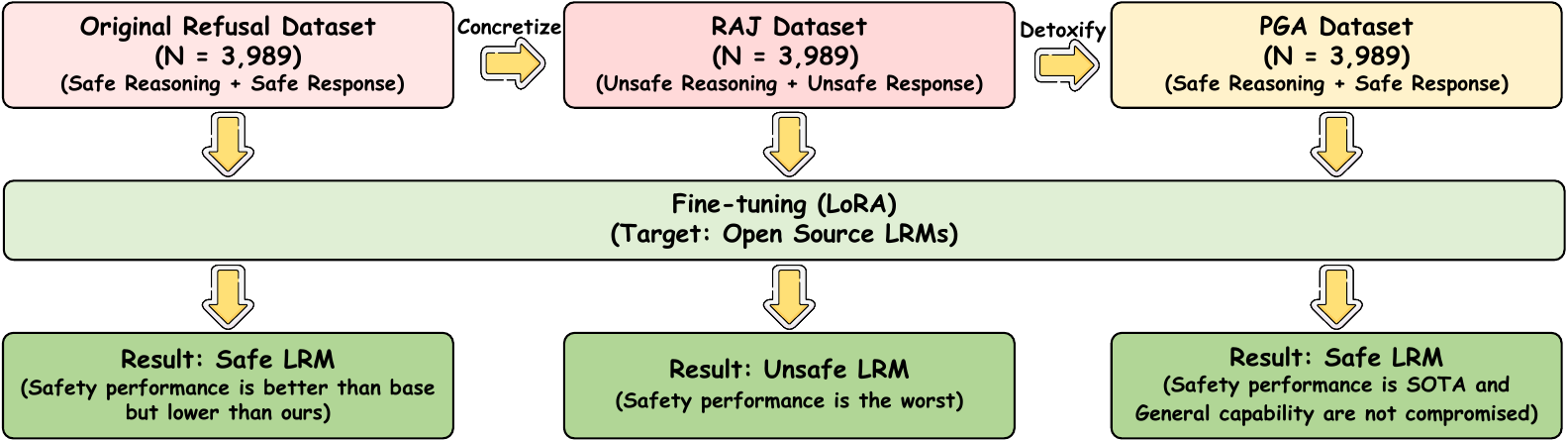}
  \caption{Ablation setup for RAJ-PGA framework. Models are fine-tuned on three datasets (Original-Refusal, RAJ, and PGA) and evaluated on safety and general benchmarks.}
  \label{fig:ablation}
\end{figure}

To rigorously verify the contribution of RAJ-PGA framework, we conducted an ablation study comparing PGA against two strategic variants (Original-Refusal and RAJ). The variants are defined as shown in Fig.~\ref{fig:ablation} and discussed as follows:

\textbf{Original-Refusal}: This variant comprises the original malicious prompts and their corresponding safe responses generated by the victim model before concretization. This baseline represents a standard refusal-based alignment strategy, serving to determine whether RAJ-PGA offers superior trade-offs compared to simply learning from direct refusals.

\textbf{RAJ}: This variant utilizes the raw RAJ dataset without alignment. Evaluating RAJ quantifies the inherent risk of the elicited capabilities and validates the necessity of the PGA.

\begin{table*}[!hbtp]
\caption{Combined Ablation Study: Safety, General Performance, Sum, and F1 Score (in \%). We compare the models fine-tuned on the Original-Refusal, the RAJ, and PGA dataset. \textbf{Bold} indicates the best result within each model subset.}
\label{tab:combined-ablation-final-bold}
\centering
\adjustbox{max width=\textwidth}{
\begin{tabular}{l | c c c c c c | c c c c c c | cc}
\toprule
 & \multicolumn{6}{c|}{\textbf{Safety Performance}} & \multicolumn{6}{c|}{\textbf{General Performance}} & \multicolumn{2}{c}{\textbf{Overall}} \\
\cmidrule(lr){2-7} \cmidrule(lr){8-13} \cmidrule(lr){14-15}
\textbf{Model} & \textbf{WildChat} & \textbf{JBB} & \textbf{StrongReject} & \textbf{WildJailbreak} & \textbf{AdvBench} & \textbf{Avg.} & \textbf{HumanEval} & \textbf{Math500} & \textbf{ARC} & \textbf{TruthfulQA} & \textbf{GPQA} & \textbf{Avg.} & \textbf{Avg.} & \textbf{F1} \\
\midrule
\rowcolor{gray!10} \multicolumn{15}{l}{\textbf{DeepSeek-R1-0528-Qwen3-8B Subset}} \\
R1-Qwen3-8B & 94.00 & 74.00 & 72.80 & 81.20 & 73.65 & 79.13 & 48.78 & \textbf{71.00} & 78.26 & 44.01 & \textbf{73.23} & 63.06 & 71.09 & 70.18 \\
\textbf{~~-PGA} & \textbf{97.00} & \textbf{83.00} & \textbf{92.40} & \textbf{96.00} & \textbf{99.62} & \textbf{93.60} & \textbf{55.49} & 65.50 & 88.29 & \textbf{73.83} & 72.73 & \textbf{71.17} & \textbf{82.38} & \textbf{80.86} \\
~~-Original-Refusal & 93.50 & 63.67 & 80.00 & 84.40 & 73.65 & 79.04 & 54.88 & 69.50 & \textbf{88.96} & 64.62 & 71.21 & 69.83 & 74.44 & 74.15 \\
~~-RAJ & 94.00 & 57.33 & 74.80 & 81.20 & 62.50 & 73.97 & 54.27 & 67.50 & 85.28 & 64.91 & 67.68 & 67.93 & 70.95 & 70.82 \\
\midrule
\rowcolor{gray!10} \multicolumn{15}{l}{\textbf{DeepSeek-R1-Distill-Llama-8B Subset}} \\
R1-Llama-8B & \textbf{98.00} & 53.67 & 58.40 & 67.60 & 46.73 & 64.88 & 41.46 & \textbf{72.50} & 79.93 & 55.12 & \textbf{76.26} & \textbf{65.05} &  64.96 & 64.97 \\
\textbf{~~-PGA} & 97.50 & \textbf{82.67} & \textbf{98.80} & \textbf{96.00} & \textbf{96.92} & \textbf{94.38} & 30.49 & 69.50 & \textbf{82.94} & \textbf{63.45} & 75.25 & 64.33 & \textbf{79.35} & \textbf{76.51} \\
~~-Original-Refusal & 97.00 & 62.67 & 83.60 & 83.20 & 57.12 & 76.72 & 35.98 & 62.50 & 82.27 & 55.26 & 71.72 & 61.55 & 69.14 & 68.30 \\
~~-RAJ & 97.00 & 54.33 & 64.40 & 70.00 & 37.12 & 64.57 & \textbf{42.68} & 70.00 & 79.60 & 52.19 & 75.76 & 64.05 & 64.31 & 64.31 \\
\midrule
\rowcolor{gray!10} \multicolumn{15}{l}{\textbf{DeepSeek-R1-Distill-Qwen-7B Subset}} \\
R1-Qwen-7B & \textbf{98.00} & 56.33 & 49.60 & 60.80 & 37.12 & 60.37 & 38.41 & \textbf{75.00} & \textbf{76.25} & 43.57 & 48.99 & \textbf{56.44} & 58.40 & 58.34 \\
\textbf{~~-PGA} & \textbf{98.00} & \textbf{71.67} & \textbf{87.20} & \textbf{88.80} & \textbf{62.88} & \textbf{81.71} & 34.15 & 68.00 & 75.59 & \textbf{46.93} & 43.94 & 53.72 & \textbf{67.72} & \textbf{64.82} \\
~~-Original-Refusal & 96.50 & 63.00 & 70.00 & 73.20 & 45.58 & 69.66 & \textbf{40.85} & 72.50 & 73.58 & 46.78 & 48.48 & \textbf{56.44} & 63.05 & 62.35 \\
~~-RAJ & 96.50 & 54.33 & 48.80 & 64.00 & 24.23 & 57.57 & 40.24 & 69.50 & 68.90 & 44.30 & \textbf{50.00} & 54.59 & 56.08 & 56.04 \\
\midrule
\rowcolor{gray!10} \multicolumn{15}{l}{\textbf{Qwen3-8B Subset}} \\
Qwen3-8B & 95.50 & 59.33 & 89.20 & 60.80 & \textbf{89.04} & 78.77 & \textbf{64.02} & 79.50 & 89.97 & \textbf{70.47} & 54.04 & \textbf{71.60} & 75.19 & 75.02 \\
\textbf{~~-PGA} & \textbf{96.00} & \textbf{72.33} & \textbf{92.00} & \textbf{83.20} & 85.58 & \textbf{85.82} & 52.44 & \textbf{81.50} & \textbf{91.97} & 67.40 & \textbf{64.14} & 71.49 & \textbf{78.66} & \textbf{78.00} \\
~~-Original-Refusal & 94.50 & 57.00 & 65.20 & 64.80 & 61.73 & 68.65 & 59.76 & 75.50 & 90.64 & 60.67 & 55.05 & 68.32 & 68.49 & 68.48 \\
~~-RAJ & 94.50 & 55.67 & 53.60 & 62.80 & 69.23 & 67.16 & 59.15 & 74.50 & 89.63 & 59.94 & 48.99 & 66.44 & 66.80 & 66.80 \\
\bottomrule
\end{tabular}
}
\end{table*}

We fine-tuned the four base LRMs using these datasets and evaluated them across both safety and general benchmarks. The results are summarized in Table~\ref{tab:combined-ablation-final-bold}.

Comparing PGA with Original-Refusal reveals the advantage of our alignment strategy. While Original-Refusal provides a decent safety baseline (e.g., $76.72\%$ on R1-Llama-8B), PGA consistently achieves higher safety performance~( $94.38\%$ ) and F1 scores( $76.51\%$ ). This suggests that training on aligned reasoning traces, which explain risks and provide educational guidance, leads to more robust safety alignment than merely mimicking refusal patterns.

Comparing PGA with RAJ, we observe an improvement in safety performance. For the DeepSeek-R1-Distill-Llama-8B Subset, the average safety performance increases from $64.57\%$ (RAJ) to 94.38\%, with practically no loss in general capability ($64.05\%$ vs. $64.33\%$). This confirms that our alignment strategy successfully removes the harmful knowledge elicited by the attack.

\section{Conclusion}
In this work, we addressed the safety risks inherent in the reasoning process of LRMs by proposing and validating a novel method for constructing alignment datasets. Our method first effectively captures LRMs' deep reasoning vulnerabilities through a Reasoning-Activated Jailbreak attack, then reshapes the harmful outputs into safe and instructive content using a Principle-Guided Alignment framework. The core contribution of our work is a systematic and scalable paradigm for AI safety, moving beyond reactive defenses against specific attacks to a more proactive alignment of the model's internal processes. The effectiveness of our method is demonstrated by the PGA dataset, which, when used for fine-tuning, significantly enhances model robustness while preserving reasoning capabilities.

The dataset construction methodology we have introduced is highly generalizable. Future work could involve applying this method to bolster model resilience against evolving adversarial reasoning attacks and multi-step prompt injections. Furthermore, the framework can be adapted for more advanced future models, providing a durable path toward building inherently safer and more reliable AI systems.

\bibliographystyle{IEEEtran}
\bibliography{arxiv}

\end{document}